\documentclass[11pt]{article}

\usepackage[]{acl}

\usepackage{times}
\usepackage{latexsym}

\usepackage[T1]{fontenc}
\usepackage[utf8]{inputenc}

\usepackage{microtype}

\usepackage{inconsolata}

\usepackage{graphicx}
\usepackage{dblfloatfix}
\usepackage{placeins}
\usepackage{amssymb}
\usepackage{booktabs}
\usepackage{amsmath} 
\usepackage{longtable}
\usepackage{array}
\usepackage{supertabular}
\usepackage[table]{xcolor}
\usepackage{hyperref}
\usepackage{amssymb}
\usepackage{float}
\usepackage{fancyhdr}
\usepackage{xcolor}

\title{Thinking Like a Botanist: Challenging Multimodal Language Models with Intent-Driven Chain-of-Inquiry}

\author{
  \textbf{Syed Nazmus Sakib\textsuperscript{1,\dag}},
  \textbf{Nafiul Haque\textsuperscript{1,\dag}},
  \textbf{Shahrear Bin Amin\textsuperscript{2}},
  \textbf{Hasan Muhammad Abdullah\textsuperscript{3}},
\\
  \textbf{Md Mehedi Hasan\textsuperscript{1}},
  \textbf{Mohammad Zabed Hossain\textsuperscript{4}},
  \textbf{Shifat E. Arman\textsuperscript{1,*}}
\\
\\
  \textsuperscript{1}Department of Robotics and Mechatronics Engineering, University of Dhaka, Dhaka, Bangladesh
\\
    \textsuperscript{2}Department of Computer Science and Engineering, University of Dhaka, Dhaka, Bangladesh
\\
  \textsuperscript{3}Department of Agronomy, Gazipur Agricultural University, Gazipur, Bangladesh
\\
  \textsuperscript{4}Department of Botany, University of Dhaka, Dhaka, Bangladesh
\\
\\
   \small{
     \textsuperscript{\dag}Equal Contribution \quad
     \textsuperscript{*}Corresponding Author
   }
}

\begin{document}
\maketitle

\begin{abstract}
Vision evaluations are typically done through multi-step processes. In most contemporary fields, experts analyze images using structured, evidence-based adaptive questioning. In plant pathology, botanists inspect leaf images, identify visual cues, infer diagnostic intent, and probe further with targeted questions that adapt to species, symptoms, and severity. This structured probing is crucial for accurate disease diagnosis and treatment formulation. Yet current vision-language models are evaluated on single-turn question answering. To address this gap, we introduce PlantInquiryVQA, a benchmark for studying multi-step, intent-driven visual reasoning in botanical diagnosis. We formalize a Chain of Inquiry framework modeling diagnostic trajectories as ordered question-answer sequences conditioned on grounded visual cues and explicit epistemic intent. We release a dataset of 24,950 expert-curated plant images and 138,068 question-answer pairs annotated with visual grounding, severity labels, and domain-specific reasoning templates. Evaluations on top-tier Multimodal Large Language Models reveal that while they describe visual symptoms adequately, they struggle with safe clinical reasoning and accurate diagnosis. Importantly, structured question-guided inquiry significantly improves diagnostic correctness, reduces hallucination, and increases reasoning efficiency. We hope PlantInquiryVQA serves as a foundational benchmark in advancing research to train diagnostic agents to reason like expert botanists rather than static classifiers.
\end{abstract}

\section{Introduction}

\begin{figure}[t]
  \includegraphics[width=\columnwidth]{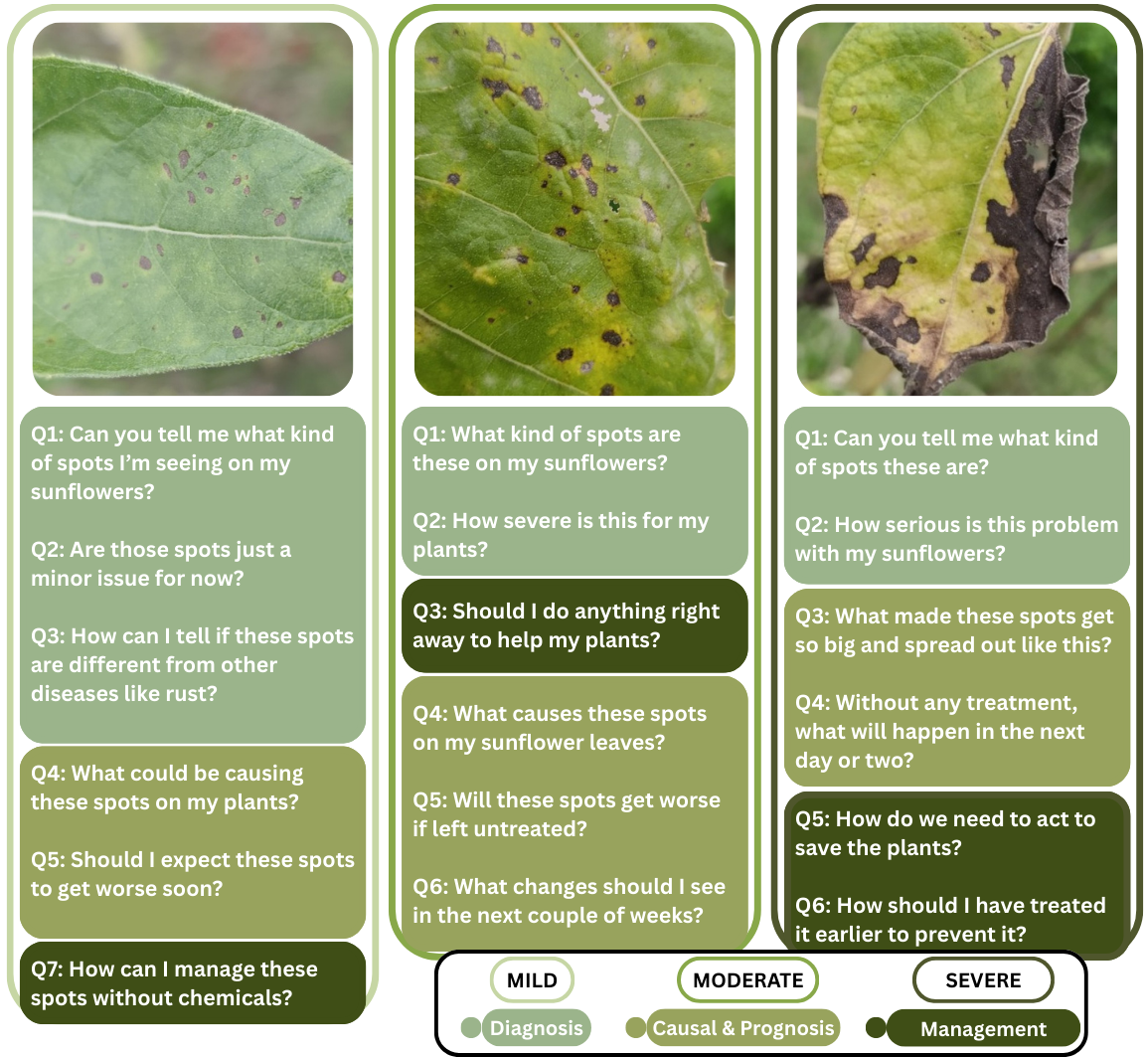}
  \caption{Unlike static QA datasets that ask generic questions regardless of disease status, our framework aligns the epistemic intent of the inquiry with visual severity. The diagram illustrates how the questioning focus evolves with disease progression in Sunflower (Alternaria Leaf Spot). The cognitive task transitions from Ambiguity Resolution (Diagnosis) in early stages to Future Forecasting (Prognosis) and Action Planning (Management) in advanced stages, ensuring questions are contextually relevant to the visual evidence.}
  \label{fig:sev-int-coi}
\end{figure}

\textbf{Visual Question Answering (VQA)} \cite{antol2015vqa} datasets have become a central paradigm for evaluating multimodal reasoning, with applications spanning medical imaging, scientific image analysis, and embodied agents \citep{das2017visual, abacha2019vqa, liu2024medcot}. Recent \textbf{VQA} benchmarks \citet{wei2022chain} have largely contributed in improving Vision Language Models' capability of understanding complex multiobject-multifocus scenarios. In \textbf{VQA}, images are analyzed in the context of a given question, requiring strong understanding of both visual cues and natural language processing \cite{antol2015vqa, andreas2016neural, abacha2019vqa}. Advanced \textbf{VQA} datasets now focus on multipanel, multichoice and strong visual-language grounded question-answer (QA) pairs \citep{fan2024muffin, liu2024medcot}. These large datasets enable \textbf{Multimodal Large Language Models (MLLMs)} to have more nuanced understanding and generation of content that blends both visual and linguistic elements. Despite these advances, most existing VQA benchmarks and plant-focused vision datasets remain fundamentally question-centric: they treat each image as an independent input to a single query or a static set of QA pairs, rather than as the starting point of a goal-directed, adaptive inquiry. In agricultural vision specifically, widely used datasets primarily target classification and segmentation tasks, such as plant disease recognition and leaf-level diagnosis, and do not capture the hierarchical, evidence-conditioned questioning strategies employed by domain experts \cite{singh2020plantdoc, mohanty2016using}.

\begin{figure*}[t]
  \includegraphics[width = 1\linewidth]{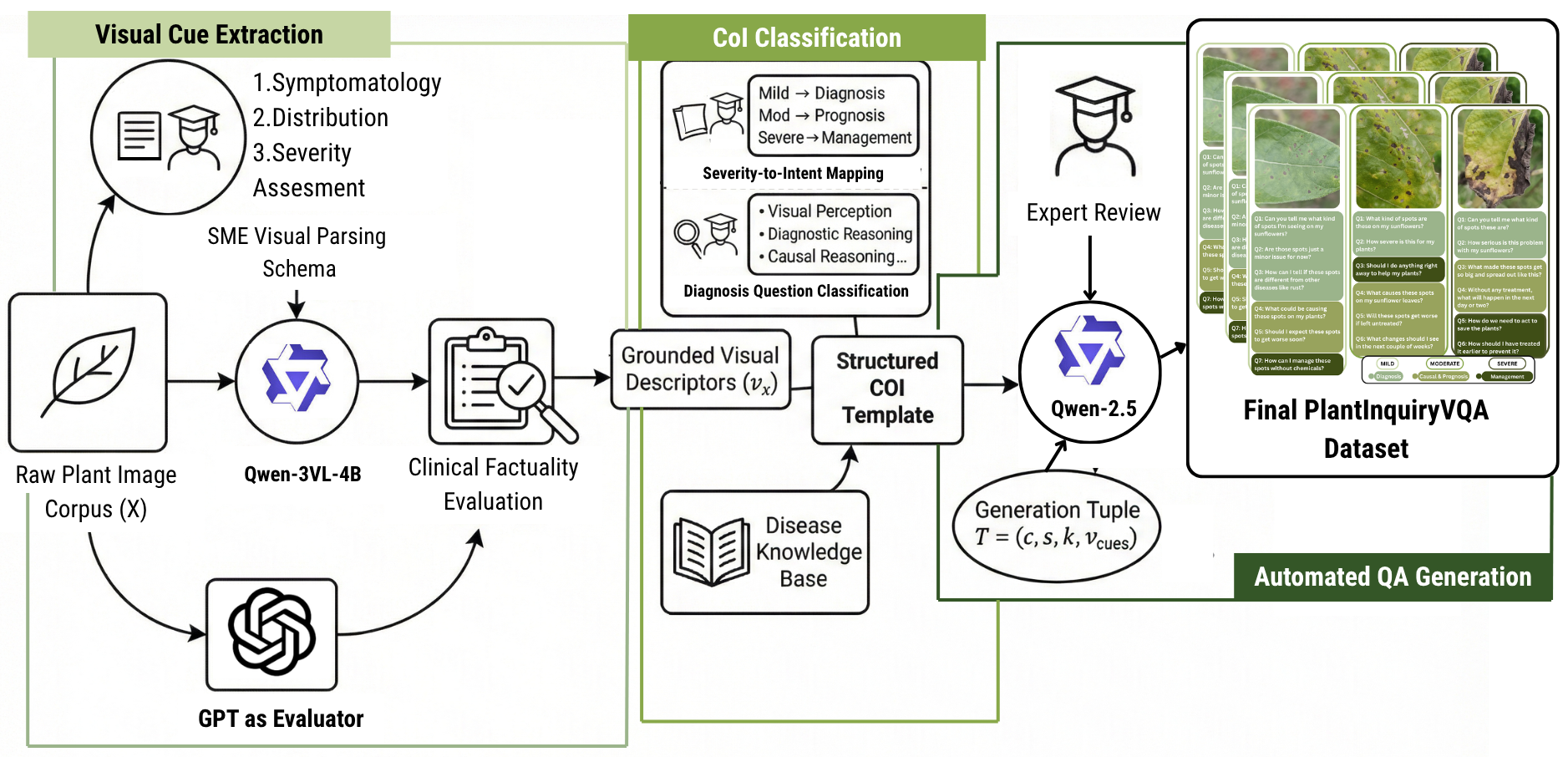}
  \caption{Overall Methodology Pipeline for PlantInquiryVQA CoI Dataset Generation. The process is divided into three phases: (1) Extracting grounded visual cues using VLM guided by expert schemas; (2) Structuring botanical knowledge to map disease severity to diagnostic intent; and (3) A dynamic LLM generation pipeline that injects specific reasoning modules based on the determined intent and visual evidence.}
\label{fig:methodology_pipeline}
\end{figure*}

However, in many real-world applications, effective visual reasoning does not arise from answering isolated questions. Instead, it emerges from a deliberate sequence of interdependent inquiries, where each question is conditioned on prior observations, and follows a sequential narrative trajectory \citep{pearl2018book, andreas2016neural}. The sequence and intent of the questions are as critical as the answers themselves. This is particularly evident in the field of botanical science where each plant sample is given unique consideration based on its visual appearance. Expert botanists conduct the holistic evaluation of a leaf sample, from species identification, to disease diagnosis and prognosis-prediction through a process of structured hierarchical and evidence-driven questioning strategy \cite{agrios2005plant, schumann1991plant}. This process, which we refer to as a \textbf{Chain-of-Inquiry} (CoI), is grounded in visual cues identified from the sample image and varies substantially depending on the plant's health condition. For example, when disease symptoms are ambiguous, experts prioritize differential diagnosis, comparative visual analysis, and disease progression prediction \cite{strange2003introduction}. Conversely, for samples exhibiting severe damage and clearly manifested symptoms, the expert CoI shifts toward disease management, prevention strategies, and earlier intervention scenarios. As a result, each plant sample is treated as a distinct case, warranting its own adaptive questioning strategy.

Though Large Language Models have made significant progress in implementing \textbf{Chain-of-Thought} (CoT) reasoning for high level planning and task completion, similar multistep exploration remains underexplored in designing VQA datasets \cite{wei2022chain, kojima2022large}. This gap raises a fundamental question: how well can similar structured probing be translated through visual Question-Answer pairs. In this paper, we introduce PlantInquiryVQA a multi-step hierarchical question answering benchmark with 24,950 curated and expert validated plant images and ~138k corresponding Question-Answer pairs related to the botanical field. We design 7 distinct question categories capable of extracting all relevant information regarding a plant sample's holistic evaluation. We then classify 12 unique visual-cue-dependent and narrative driven adaptive \emph{CoI}. Our implementation strategy allows for each plant sample to have its unique questioning trajectory specific to its visual cues with sequence of interconnected questions and answers. These serve as the precondition for in-depth multistage reasoning and contextual semantic analysis. Finally, we benchmark popular open-source and proprietary MLLMs on the PlantInquiryVQA benchmark and conduct thorough error analysis with the help of expert verification.

Our contribution includes:
\begin{enumerate}

    \item \textbf{The PlantInquiryVQA Benchmark:} We release a large-scale dataset of 25k manually curated images across diverse crop species, annotated with expert-verified visual cue descriptions and domain-specific knowledge bases.
    
    \item \textbf{The Chain-of-Inquiry (CoI) Framework:} We formalise a novel reasoning taxonomy for PlantInquiryVQA, classifying 12 unique reasoning templates into 7 distinct cognitive categories (including Etiological Reasoning, Differential Diagnosis, and Counterfactual Analysis).
    
    \item \textbf{Diagnostic Reasoning Evaluation:} We conduct a comprehensive evaluation of both closed- and open-source MLLMs. Our results show that question-guided protocols significantly reduce hallucination and improve diagnostic correctness, while sequential chains of inquiry enhance explainability efficiency. %These findings reveal that current MLLMs lack robust botanical reasoning, and underscore the importance of inquiry structure for reliable plant disease diagnosis.
    
\end{enumerate}

\section{Related Work}

\textbf{Visual Question Answering (VQA).} \citet{antol2015vqa} combines image understanding with natural language processing to answer queries about visual content. Unlike standard classification, VQA facilitates interactive question–answering, allowing models to capture and articulate complex relationships within an image. As such, the application of VQA now spans multiple domains. These include: educational tools, customer service systems, and autonomous driving etc. In particular, VQA shows exceptional potential in the field of pathological diagnosis and health inquiry \citep{zhang2023pmc}. Current medical VQA benchmarks include PMC-VQA \citet{zhang2023pmc}, SLAKE \citet{liu2021slake}, Path-VQA \citet{he2020pathvqa}, and VQA-RAD \citet{lau2018dataset}. However, these datasets are focused on medical diagnostics. In agriculture, existing popular datasets like PlantVillage \citep{Hughes2015PlantVillage}, PlantDoc \citet{singh2020plantdoc}, and PlantSeg \citet{wolny2020accurate} focus on classification or segmentation tasks. While they support disease detection, they do not enable interactive reasoning through question–answer formats.

\textbf{Chain-of-Thought in Multimodal Models}. Parallel to these advances, benchmark datasets have also evolved to test deeper cognitive capabilities through Chain-of-Thought (CoT). Originally proposed for text-based Large Language Models by  \citet{wei2022chain}, CoT marked a paradigm shift by prompting models to decompose complex problems into intermediate logical steps rather than mapping inputs directly to outputs. This methodology has recently been adapted for Multimodal Large Language Models (MLLMs), where the reasoning chain must ground linguistic tokens in visual features to reduce hallucination \cite{zhang2023multimodal}. Recent frameworks like Multimodal-CoT \citet{zhang2023multimodal, liu2024medcot} demonstrate that incorporating explicit reasoning paths significantly improves performance on complex vision-language tasks. However, in the majority of these works, CoT is treated primarily as a prompting strategy or a latent capability of the model architecture, rather than an explicit structural requirement inherent to the dataset itself.

\textbf{Structured and Hierarchical VQA}. Consequently, there have been efforts to translate this step-by-step reasoning into VQA dataset design. Recent benchmarks have attempted to introduce structure into visual questioning; for instance, BloomVQA \citet{gong2024bloomvqa} organizes questions based on Bloom’s taxonomy of cognitive complexity, while MedCoT \citet{liu2024medcot} utilizes hierarchical expert agents to simulate medical reasoning flows. Similarly, conversational benchmarks like \citet{das2017visual} introduced the concept of multi-turn visual dialogue. However, most existing hierarchical datasets rely on static question taxonomies or crowd-sourced dialogues that lack the goal-oriented precision of a domain expert \cite{gong2024bloomvqa}. They generally fail to capture the causal dependency of professional diagnosis, where the formulation of the next question is strictly conditional on the visual evidence verified in the previous step. \textbf{PlantInquiryVQA} fills this gap by making the CoI, an explicit dataset-level artifact: the dataset’s question sequences are constructed to mirror the adaptive, decision-driven workflows of domain experts.

\section{Methodology}

\subsection{Formalization of Chain-of-Inquiry}
\label{sec:coi_form}
We define the \textbf{CoI} as a visual-semantic trajectory conditioned on diagnostic intent. Let $\mathcal{X}$ denote the set of plant images and $\mathcal{V}$ be the space of explicit visual cues (e.g., \textit{chlorosis patterns}, \textit{lesion margins}). For a given image $x \in \mathcal{X}$, we extract a set of grounded visual descriptors $v_x \in \mathcal{V}$.

CoI $C$ is defined as an ordered sequence of $T$ dialogue turns:
\begin{equation}
    C(x, v_x) = \big\langle (q_1, a_1), (q_2, a_2), \dots, (q_T, a_T) \big\rangle
\end{equation}
where each question $q_t$ is conditioned on the visual evidence $v_x$, the previous context $H_{t-1} = \{(q_i, a_i)\}_{i=1}^{t-1}$, and a latent diagnostic intent $k \in \mathcal{K}$. Here, the intent space $\mathcal{K}$ is stratified into three primary epistemic goals derived from expert botany:

1. Diagnosis ($k_D$): Identifying health status and discriminating between similar pathologies (Differential Diagnosis).

2. Prognosis ($k_P$): Predicting disease trajectory, temporal evolution, and causal etiology.

3. Management ($k_M$): Prescribing remediation strategies and formulating counterfactual preventative reasoning.

Thus, the generation of a specific CoI is modeled as sampling from a conditional distribution $P(C \mid x, v_x, k)$, ensuring that the dialogue trajectory aligns with the clinical necessity of the plant sample.

 \subsection{Visual Cue Extraction and CoI Classification}
 In extracting relevant visual cues from a sample image, we strictly adhere to the \textbf{Symptomatological Diagnostic Protocol}, as defined in \citet{agrios2005plant}. We recruited two PhD-level and four graduate-level botanists specializing in plant pathology as Subject Matter Experts (SMEs) to define a "Visual Parsing Schema". Following the diagnostic criteria outlined by \citet{agrios2005plant, streets1972diagnosis}, the experts established a structured feature extraction template comprising three critical diagnostic dimensions: \emph{Symptomatology}, \emph{Distribution Patterns} and \emph{Disease Severity Quantification}. A comprehensive description of these criteria are provided in the appendix \ref{sec:expert_schema}. In the pilot phase, each SME applied this schema to a randomized batch of 50 images. To ensure inter-annotator consistency, batches were cross-verified among the group, synthesizing a unified, expert-validated template.
\paragraph{Automated Extraction \& Validation.} Leveraging this expert-derived schema, we prompted three open-source Vision-Language Models (VLMs) to generate dense, fine-grained visual cue descriptions  for a batch of 250 images. As shown by the comparative benchmarking in \autoref{tab:ablation}, \textbf{Qwen3-VL-4B} outperformed all other models. Consequently, we used \textbf{Qwen3-VL-4B} to extract visual cues for the entire corpus of 24,950 images. Given the scale of the dataset, exhaustive expert annotation was infeasible. We therefore employed a hybrid validation strategy: (1) Qwen3-VL extracted visual cues using expert-designed schemas; (2) \textbf{GPT-4V} \citet{achiam2023gpt} acted as an external evaluator to cross-verify the extracted cues, flagging outputs with high semantic divergence from Qwen3-VL; (3) finally, domain experts conducted a rigorous \textbf{Clinical Factuality} check on all flagged instances plus 5,000 randomly sampled images. Experts annotated errors in two categories: \textit{Object Hallucination} (mentioning symptoms, e.g., "halo", not present in the image) and \textit{Attribute Mismatch} (incorrectly describing color or texture). The model achieved a Factuality Score of 93.8\%, defined as the proportion of generations free from critical clinical errors. Appendix \ref{sec:vis_cus_example} includes the comparative analysis of model–extracted visual cues together with the expert-annotated, knowledge-based cues.

We rely on human specialists to collect high-quality groundtruth QA chains for plant pathology diagnosis. We ask our botanists to compile relevant and semantically rich question samples from established botanical sources \citep{agrios2005plant, schumann1991plant, strange2003introduction, streets1972diagnosis}. While classical literature outlines the biological phases of diagnosis (e.g., Symptomatology, Etiology, Epidemiology), there is no standard taxonomy for interrogating them in a visual dialogue. To bridge this gap, our SMEs classify the standard diagnostic inquiries in 7 general categories: \textbf{Visual Perception \& Grounding:} Corresponds to the \textit{Symptomatology} \citep{strange2003introduction}. \textbf{Diagnostic Reasoning:} Aligns with the \textit{Differential Diagnosis} \citep{streets1972diagnosis}. \textbf{Causal Reasoning:} Derived from \textit{Etiology}. \textbf{Risk Assessment:} Maps to \textit{Epidemiology}. \textbf{Prognosis Prediction:} Models the \textit{Disease Cycle}. \textbf{Prescriptive Reasoning:} Corresponds to \textit{Disease Management}. \textbf{Counterfactual Reasoning:} Simulates retrospective analysis (e.g., `What if treatment had occurred earlier?'), a crucial component of post-epidemic evaluation and learning. 

\begin{figure*}[t]
  \includegraphics[width = 1\linewidth]{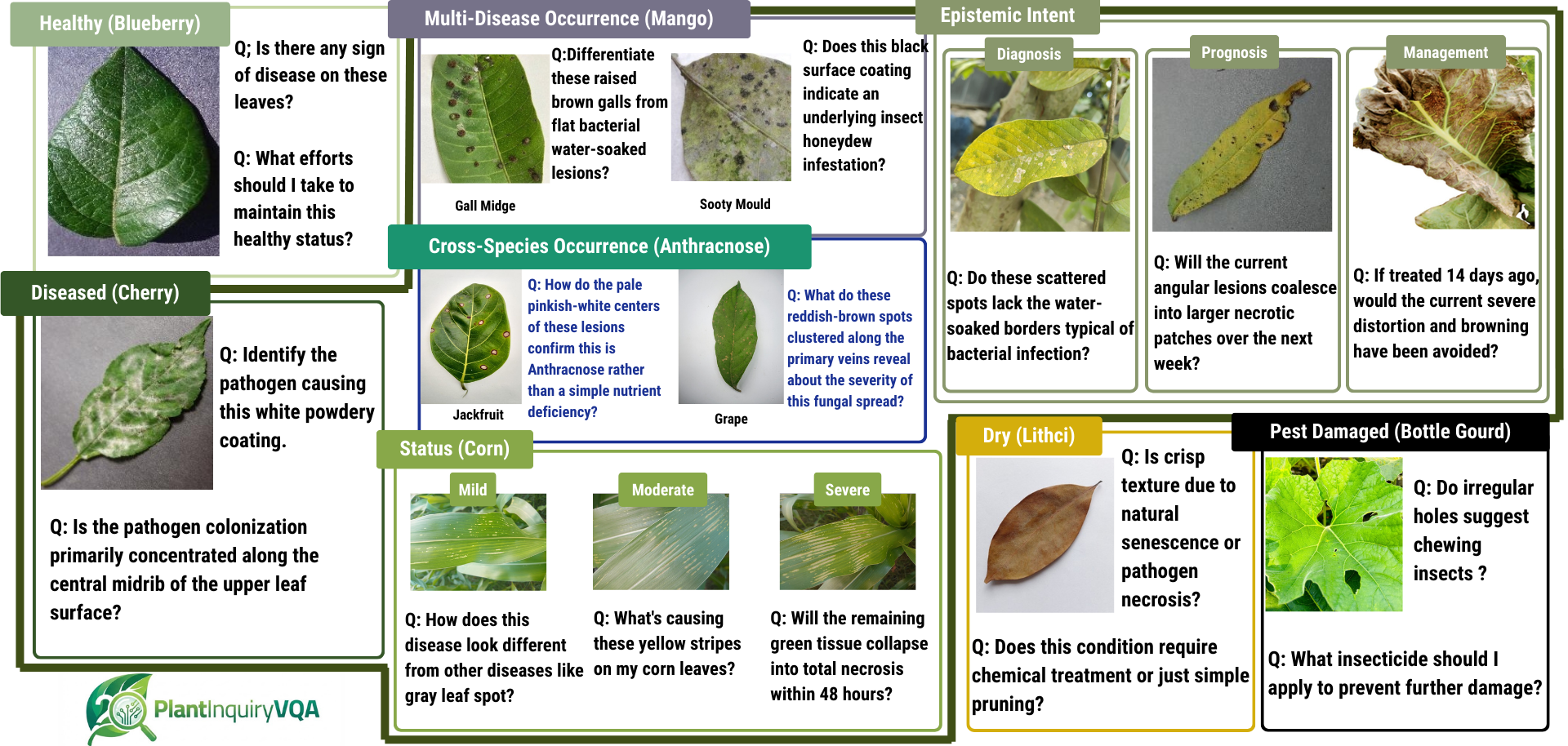} 
  \caption {Qualitative Examples of 12 Distinct CoI Trajectories. The framework adapts questioning strategies across four varying axes of diagnostic complexity: (1) Health Status (Healthy, Diseased, Senescence, Pest Damaged); (2) Disease Severity, showing evolving logic for Mild, Moderate, and Severe Maize infection; (3) Instance Variety, handling Multi-disease and Cross-species constraints; and (4) Epistemic Intent, illustrating the cognitive shift from Diagnosis to Prognosis and Management based on visual evidence.}
  \label{fig:coi_traj}
\end{figure*}

 We choose a random sample of 600 images from our dataset and ask the experienced botanists to conduct a clinical evaluation of each sample and note down their questioning strategy. Here we identify three distinct intents evident among the questioning arc for a given plant image. These intents are explained in \autoref{sec:coi_form} and further explained here: 
 For samples exhibiting \textbf{Mild} or ambiguous symptoms, the CoI intent \textbf{Diagnosis} ($k_D$) focuses on \textit{differential diagnosis}. Botanists compare disease symptoms and ask clarifying questions to distinguish the observed pathology from potential lookalikes. Conversely, for samples classified as \textbf{Moderate}, the focus shifts to \textbf{Causal \& Prognosis} ($k_P$). Here, the inquiries emphasize understanding the disease's spread and inferring the environmental conditions such as humidity or soil pH that likely facilitated the infection. Finally, for \textbf{Severe} cases where damage is extensive, the expert intent shifts to \textbf{Management} ($k_M$). The reasoning chain here is primarily focused on identifying immediate strategies for disease remedy and employing counterfactual analysis (e.g., \textit{"How would the plant's condition differ if intervention had occurred two weeks prior?"}) to simulate critical care scenarios. Following this schema, we translate the botanist chain of thought into 12 distinct \textbf{Chain-of-Inquiry}, covering \textit{4 Health Status (Healthy, Diseased, Senescence, Pest Damaged), 3 Disease Severity (Mild, Moderate, Severe), 2 Instance Variety (Multi-disease Occurrence, Cross-species Occurrence) and 3 Epistemic Intent (Diagnosis, Prognosis, Management)}. Appendix \ref{sec:coi_traj} shows each \textbf{CoI} with corresponding image and varying question samples.

\begin{table}
  \centering
  \begin{tabular}{lc}
    \hline
    \textbf{Model} & \textbf{Score (\%)} \\
    \hline
    Qwen3-VL   & 184 (73.6\%) \\
    Gemma3-4B  & 58 (23.2\%)  \\
    LLaVA-1.5  & 8 (3.2\%)    \\
    \hline
  \end{tabular}
  \begin{tabular}{lc}
    \hline
    \textbf{Model} & \textbf{Score (\%)} \\
    \hline
    Qwen3-VL   & 185 (74.0\%) \\
    Gemma3-4B  & 61 (24.4\%)  \\
    LLaVA-1.5  & 4 (1.6\%)    \\
    \hline
  \end{tabular}
  \caption{Comparative analysis of three open-source models for visual grounding (top) and specificity (bottom).}
  \label{tab:ablation}
\end{table}

\subsection{Dataset Generation Pipeline}
\label{sec:dataset_gen}
We developed a structured generation pipeline governed by the configuration tuple $T=(c,s,k_s,V_{cues})$, corresponding to \textit{Biological condition, Severity, Severity derived Intent}, and \textit{Visual cues}. This decoupling allows us to generate diverse reasoning chains even for the same image (e.g., asking for \textit{Management} advice on a \textit{Mild} case vs. a \textit{Severe} case). Here, we employed Qwen2.5-7B-Instruct \citep{yang2025qwen2} to dynamically assemble dialogue trajectories from question templates. Here, the epistemic goal k (\textit{Diagnosis} ($k_D$), \textit{Prognosis} ($k_P$) and \textit{Management}($k_M$)) modulates the information density based on severity status $s$. To further enhance complexity, we inject specific reasoning modules (e.g., \texttt{temporal\_evolution}, \texttt{remediation\_strategy} etc.) directly into the chains. \autoref{fig:methodology_pipeline} shows the overall generation of the \textbf{PlantInquiryVQA} dataset. Our tiered approach ensures that PlantInquiryVQA covers the full spectrum of diagnostic difficulty, from routine identification to complex, multi-step clinical reasoning. We provide the detailed module injection logic and generation protocols in Appendix \ref{sec:code}.

\section{Experiment}

\begin{table*}[t]
\centering
\small
\setlength{\tabcolsep}{5pt}
\begin{tabular}{l ccc ccccc cc}
\toprule
& \multicolumn{3}{c}{\textbf{Lexical Metrics}} & \multicolumn{5}{c}{\textbf{Domain Alignment \& Quality}} & \multicolumn{2}{c}{\textbf{Fairness}} \\
\cmidrule(lr){2-4} \cmidrule(lr){5-9} \cmidrule(lr){10-11}
\textbf{Model} & \textbf{F1} & \textbf{BLEU-4} & \textbf{R-L} & \textbf{Dis} & \textbf{Clin.} & \textbf{Safe.} & \textbf{VG} & \textbf{Len.} & \textbf{B} & \textbf{F} \\
\midrule
\textsc{Gemini-3-Flash}        & \textbf{0.255} & \textbf{0.033} & \textbf{0.196} & \textbf{0.444} & \textbf{0.188} & \textbf{0.147} & 0.259 & 85.8  & $0.700$ & $-0.020$ \\
\textsc{Gemini-2.5-Pro}        & 0.225 & 0.016 & 0.132 & 0.357 & 0.112 & 0.040 & 0.408 & 142.9 & $0.450$ & $+0.008$ \\
\textsc{Qwen3-VL-235B}         & 0.210 & 0.013 & 0.120 & 0.348 & 0.111 & 0.035 & 0.489 & 143.9 & $0.333$ & $+0.413$ \\
\textsc{Seed-1.6-Flash}        & 0.226 & 0.022 & 0.139 & 0.344 & 0.120 & 0.075 & 0.394 & 99.1  & $0.591$ & $-0.209$ \\
\textsc{Llama-3.2-90B-Vision}  & 0.212 & 0.014 & 0.105 & 0.340 & 0.185 & 0.214 & 0.372 & 134.9 & $0.595$ & $-0.253$ \\
\textsc{Llama-4-Maverick}      & 0.212 & 0.013 & 0.103 & 0.329 & 0.175 & 0.202 & 0.397 & 144.5 & $0.500$ & $-0.564$ \\
\textsc{Gemini-2.5-Flash}      & 0.226 & 0.018 & 0.145 & 0.299 & 0.098 & 0.046 & 0.392 & 163.5 & $0.455$ & $+0.455$ \\
\textsc{Qwen3-VL-32B}          & 0.182 & 0.011 & 0.096 & 0.288 & 0.096 & 0.035 & 0.475 & 227.8 & $0.250$ & $-0.075$ \\
\textsc{Qwen-VL-Plus}          & 0.166 & 0.009 & 0.074 & 0.316 & 0.101 & 0.032 & \textbf{0.508} & 287.9 & $0.389$ & $+0.162$ \\
\textsc{Gemma-3-27B}           & 0.192 & 0.011 & 0.103 & 0.272 & 0.086 & 0.032 & 0.353 & 156.9 & 0.358 & $-0.382$ \\
\textsc{Pixtral-12B}           & 0.225 & 0.016 & 0.122 & 0.272 & 0.145 & 0.159 & 0.368 & 98.0  & $0.447$ & $-0.392$ \\
\textsc{Qwen2.5-VL-32B}        & 0.177 & 0.009 & 0.076 & 0.254 & 0.078 & 0.017 & 0.463 & 260.4 & $0.308$ & $-0.048$ \\
\textsc{Phi-4-Multimodal}      & 0.177 & 0.010 & 0.097 & 0.254 & 0.087 & 0.040 & 0.358 & 167.2 & $0.333$ & $-0.461$ \\
\textsc{Qwen2.5-VL-72B}        & 0.236 & 0.016 & 0.123 & 0.247 & 0.080 & 0.040 & 0.375 & 106.2 & $0.500$ & $+0.345$ \\
\textsc{Grok-4.1-Fast}         & 0.203 & 0.016 & 0.132 & 0.224 & 0.067 & 0.009 & \textbf{0.498} & 100.7 & $0.489$ & $-0.092$ \\
\textsc{Mistral-Medium-3.1}    & 0.211 & 0.015 & 0.119 & 0.205 & 0.062 & 0.023 & 0.360 & 110.7 & $0.536$ & $-0.352$ \\
\textsc{Ministral-8B}          & 0.180 & 0.010 & 0.094 & 0.197 & 0.060 & 0.020 & 0.394 & 151.8 & $0.542$ & $-0.149$ \\
\textsc{Ministral-3B}          & 0.166 & 0.007 & 0.083 & 0.189 & 0.059 & 0.020 & 0.372 & 163.0 & $0.440$ & $-0.088$ \\
\bottomrule
\end{tabular}
\caption{\textbf{Main Results on Test Set.} \textsc{Gemini-3-Flash} leads across lexical and domain-specific metrics with near-zero Cross-Class Fairness ($F{=}-0.020$), though its $B{=}0.700$ indicates that its residual errors still fall back to prevalent pathologies. \textsc{Seed-1.6-Flash} is competitive on disease accuracy ($0.344$) with negligible minority penalty. \textsc{Qwen3-VL-235B} and \textsc{Gemini-2.5-Flash} exhibit substantial minority penalties ($F>0.4$), mirroring long-tailed clinical failure. \textsc{Grok-4.1-Fast} retains the highest Visual Grounding ($0.498$) at balanced $B$, underscoring the grounding-vs-reasoning trade-off.}
\label{tab:combined_results}
\end{table*}

We first evaluate eighteen leading Multimodal Large Language Models (MLLMs) on the PlantInquiryVQA benchmark. Then, based on the evaluation results, we conduct a thorough error analysis and ablation study to isolate the impact of our CoI framework on diagnostic reasoning.

\subsection{Experimental Setup}
\textbf{MLLMs} The MLLMs we adopt in the evaluation include both open-weight models and proprietary models accessed via API. The open-weight MLLMs include the \textbf{Qwen} series (Qwen3-VL-32B, Qwen2.5-VL-72B/32B, Qwen-VL-Plus) , \textbf{Gemma} variants (Gemma-3-27B) , and the \textbf{Ministral} family (Ministral-8B/3B). We also evaluate a distilled lightweight model, \textbf{Nemotron-Nano-12B}, to assess performance at the edge. For proprietary models, we evaluate \textbf{Gemini-2.5-Pro}, \textbf{Gemini-2.5-Flash}, and the recently released \textbf{Gemini-3-Flash}.

\subsection{Evaluation Metrics.} Beyond standard lexical metrics (F1, BLEU-4, ROUGE-L), we introduce seven domain-specific scores to probe clinical reliability; formal definitions are deferred to Appendix~\ref{sec:metric_definitions}. 
\paragraph{Disease Identification} ($S_{dis}$) captures strict semantic retrieval of the correct pathogen. 
\paragraph{Safety} ($S_{safe}$) penalises the \textit{False Reassurance} failure mode in which a diseased sample is misclassified as healthy, the most consequential error in phytopathology.
\paragraph{Clinical Utility} ($S_{clin}$) is a composite score that aggregates identification accuracy and actionable remediation advice, discounted by safety violations; we set $(\alpha{,}\beta{,}\gamma){=}(0.5,0.3,0.2)$ consistent with standard phytopathology practice, in which pathogen identification is the prerequisite for subsequent action~\citep{agrios2005plant}, and confirm that relative model rankings are stable to substantial weight perturbation (Spearman $\rho_s > 0.91$; Appendix~\ref{sec:weight_sens}). 
\paragraph{Visual Grounding} ($S_{vg}$) reports recall over expert-verified visual cues, 
\paragraph{Visual Feature Extraction Efficiency} ($E$) measures grounded cues per 100 generated words, rewarding concise evidence over verbose filler. \\
Finally, motivated by recent holistic VLM evaluation frameworks~\citep{lee2024vhelm, zhao2025spd}, we introduce two distributional metrics: 
\paragraph{Prevalence Bias} ($B$), the fraction of misdiagnoses that default to a more common pathology than the true class.
\paragraph{Cross-Class Fairness} ($F$), the Clinical Utility gap between majority and minority crop strata, with $|F|$ near zero indicating equitable performance.

\subsection{Main Result}
We evaluate the dataset using a Cumulative Context Test, in which each successive question is conditioned on the full history of preceding questions and generated answers in the chain. Comparative performance across models is reported in \autoref{tab:combined_results}. The results reveal that \textbf{Gemini-3-Flash} model consistently outperforms all other evaluated architectures, establishing a highest score on this benchmark. It achieves the highest scores across both standard lexical metrics and domain specific alignment score. This performance gap suggests that advanced closed-source models possess superior instruction-following capabilities, allowing them to better handle the long context dependencies required by CoI. Unlike smaller models that often lose track of the diagnostic "narrative" after several turns, Gemini-3-Flash maintains coherence, accurately translating visual symptoms into precise disease identifications and safety-compliant advice.

Interestingly, while Gemini-3-Flash excels in clinical reasoning, the \textbf{Grok-4.1-Fast} maintains the highest Visual Grounding (VG) score. This indicates that while these model excel at "accurately describing raw visual features such as "yellow spots" or "necrotic margins", they struggle to synthesize that evidence into a coherent clinical diagnosis. This discrepancy highlights that "seeing" the symptom is not equivalent to "diagnosing" the pathology, underscoring the necessity of our \textbf{CoI} framework to bridge this cognitive lapse.

The most prominent finding from \autoref{tab:combined_results} is the substantial domain gap in current state-of-the-art MLLMs regarding botanical pathology. Even the top-performing model, Gemini-3-Flash, achieves a Clinical Utility score of only 0.188 and a Disease Identification score of 0.444. This indicates that PlantInquiryVQA represents a significantly hard benchmark; models struggle to translate visual signals into accurate, safe clinical diagnoses.

\subsection{Error Analysis}
Intending to identify why models fail or succeed within our framework, we move beyond simple accuracy metrics. We leverage the structured nature of PlantInquiryVQA to conduct comparative experiments that isolate specific reasoning capabilities. \\
\textbf{How does the structure of inquiry influence diagnostic accuracy?}  To evaluate if the questions themselves act as effective attention mechanisms, we conducted a \textbf{Protocol Structure Benefit Test}.

\begin{figure}[H]
  \centering
  \includegraphics[width=0.75\linewidth]{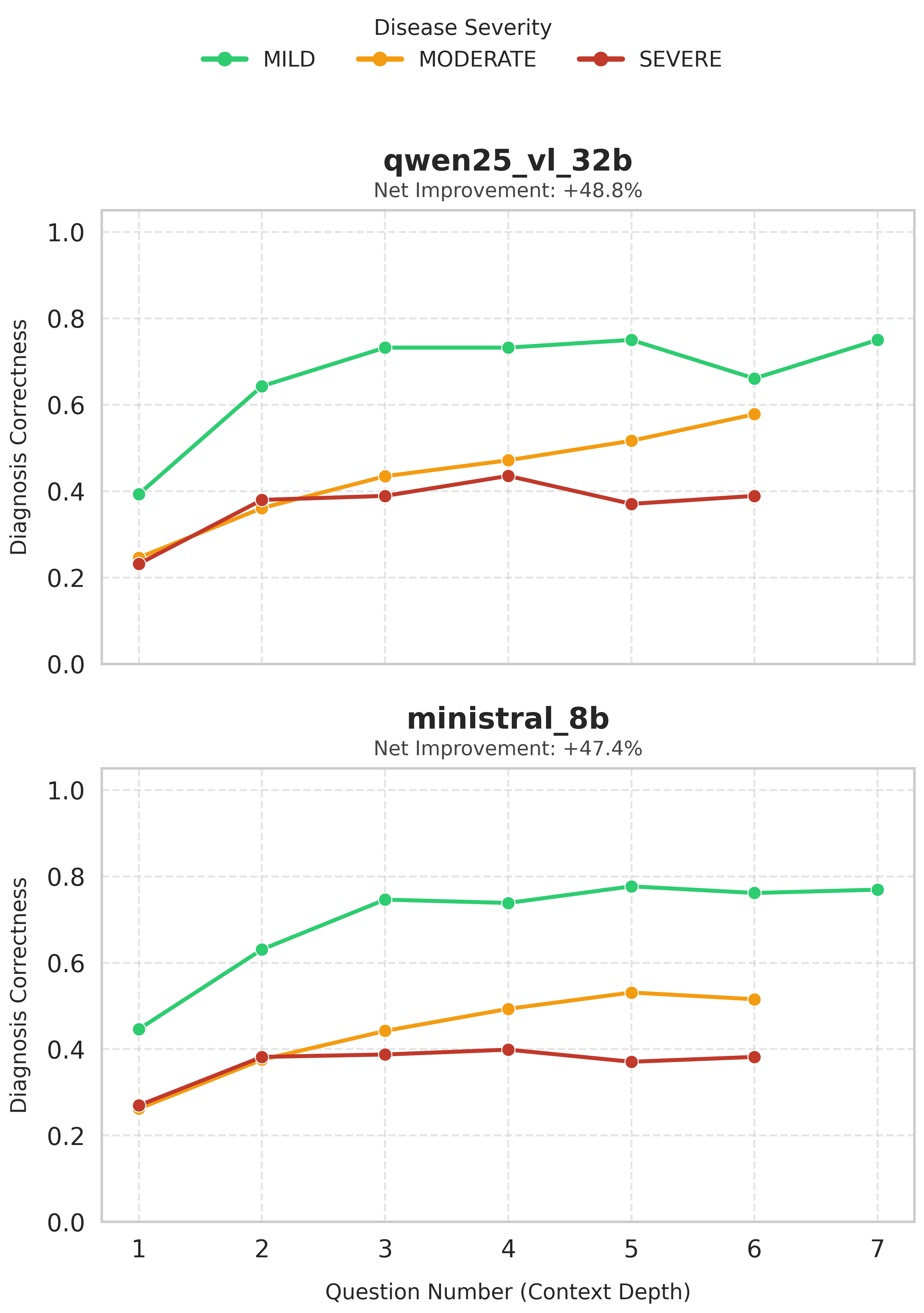}
  \caption{Protocol Structure Benefit Test for Qwen25-VL-32B and Ministral-8B. Both models achieve better Diagnosis Correctness scores using questions as context (48.8\% and 47.4\%, respectively).}
  \label{fig:diag_correct}
\end{figure}

\begin{table*}[t]
\centering
\footnotesize
\setlength{\tabcolsep}{3pt}
\begin{tabular}{l cccc cc cccc c}
\toprule
\textbf{Model}
& \multicolumn{4}{c}{\textbf{Scaffolded}}
& \multicolumn{2}{c}{\textbf{Cascading}}
& \multicolumn{4}{c}{\textbf{Guided}}
& $\Delta_{E}\%$ \\
\cmidrule(lr){2-5} \cmidrule(lr){6-7} \cmidrule(lr){8-11}
& Eff. & Cues & Words & DA
& Eff. & DA
& Eff. & Cues & Words & DA
& \\
\midrule
\textsc{Gemini-3-Flash}        & 2.60 & \textit{4.07} & \textit{186.4} & 0.264 & 3.54 & 0.347 & \cellcolor{green!25}3.67 & \textit{3.20} & \textit{108.2} & 0.444 & $+41.15$ \\

\textsc{Gemini-2.5-Pro}        & 2.95 & 6.11 & 268.6 & 0.247 & 3.45 & 0.289 & \cellcolor{green!25}3.58 & 4.00 & 147.9 & 0.357 & $+21.36$ \\

\textsc{Qwen3-VL-235B}         & 2.88 & \textit{6.45} & \textit{269.3} & 0.241 & 3.21 & 0.282 & \cellcolor{green!25}3.33 & \textit{3.95} & \textit{138.4} & 0.348 & $+15.63$ \\
\textsc{Seed-1.6-Flash}        & 3.22 & 5.63 & 198.1 & 0.256 & 3.62 & 0.299 & \cellcolor{green!25}3.75 & 4.10 & 125.0 & 0.344 & $+16.46$ \\
\textsc{Llama-3.2-90B-Vision}  & 2.40 & 4.18 & 197.6 & 0.252 & 2.75 & 0.265 & \cellcolor{green!25}2.85 & 3.50 & 140.0 & 0.340 & $+18.75$ \\
\textsc{Llama-4-Maverick}      & 2.31 & 4.32 & 208.7 & 0.229 & 2.56 & 0.267 & \cellcolor{green!25}2.65 & 3.40 & 145.0 & 0.329 & $+14.72$ \\
\textsc{Gemini-2.5-Flash}      & 2.60 & 8.65 & 456.0 & 0.207 & 3.54 & 0.242 & \cellcolor{green!25}3.67 & 4.71 & 181.0 & 0.299 & $+41.15$ \\
\textsc{Qwen3-VL-32B}          & 2.88 & 8.82 & 362.6 & 0.199 & 3.21 & 0.233 & \cellcolor{green!25}3.33 & 5.96 & 206.4 & 0.288 & $+15.63$ \\
\textsc{Gemma-3-27B}           & 1.88 & 6.61 & 411.0 & 0.188 & 2.30 & 0.220 & \cellcolor{green!25}2.38 & 4.77 & 256.2 & 0.272 & $+26.60$ \\
\textsc{Pixtral-12B}           & 2.53 & 4.00 & 217.1 & 0.188 & 2.80 & 0.220 & \cellcolor{green!25}2.90 & 3.20 & 145.0 & 0.272 & $+14.62$ \\
\textsc{Qwen2.5-VL-32B}        & 1.60 & 6.81 & 471.4 & 0.176 & 2.84 & 0.206 & \cellcolor{green!25}2.94 & 4.47 & 249.2 & 0.254 & $+83.75$ \\
\textsc{Phi-4-Multimodal}      & 1.94 & 6.71 & 423.4 & 0.176 & 2.46 & 0.206 & \cellcolor{green!25}2.55 & 4.80 & 240.0 & 0.254 & $+31.44$ \\
\textsc{Qwen2.5-VL-72B}        & 2.46 & 4.53 & 212.0 & 0.171 & 2.82 & 0.200 & \cellcolor{green!25}2.92 & 3.47 & 171.9 & 0.247 & $+18.70$ \\
\textsc{Grok-4.1-Fast}$^{*}$   & 4.54 & 8.49 & 220.6 & 0.155 & 5.02 & 0.181 & \cellcolor{green!25}5.20 & 5.80 & 135.0 & 0.224 & $+14.54$ \\
\textsc{Qwen-VL-Plus}          & 1.63 & 7.85 & 536.5 & 0.149 & 2.44 & 0.174 & \cellcolor{green!25}2.53 & 4.92 & 282.0 & 0.215 & $+55.21$ \\
% \textsc{Nemotron-Nano-12B}     & \cellcolor{green!25}3.84 & 5.72 & 164.7 & 0.146 & 3.22 & 0.170 & 3.34 & 4.32 & 165.4 & 0.210 & $-13.02$ \\
\textsc{Mistral-Medium-3.1}    & 2.35 & 4.75 & 228.6 & 0.142 & 2.61 & 0.166 & \cellcolor{green!25}2.70 & 3.60 & 155.0 & 0.205 & $+14.89$ \\
\textsc{Ministral-8B}          & 2.21 & 5.19 & 264.2 & 0.137 & 2.56 & 0.160 & \cellcolor{green!25}2.65 & 3.70 & 172.4 & 0.197 & $+19.91$ \\
\textsc{Ministral-3B}          & 2.26 & 5.30 & 247.8 & 0.131 & 2.61 & 0.153 & \cellcolor{green!25}2.71 & 3.30 & 152.1 & 0.189 & $+19.91$ \\
\midrule
\textit{Average}               & \textit{2.58} & & & \textit{0.193} & \textit{3.03} & \textit{0.227} & \textit{3.15} & & & \textit{0.278} & \textit{$+22.09$} \\
\bottomrule
\end{tabular}
\caption{\textbf{Efficiency and Diagnostic Accuracy across three context conditions.} \textit{Scaffolded}: each question in isolation; \textit{Cascading}: model's own prior answers as history (auto-regressive deployment); \textit{Guided}: ground-truth history. \textit{Eff.}~=~Efficiency Score (\S\ref{sec:metric_definitions}), \textit{DA}~=~Diagnostic Accuracy ($S_{dis}$). Green cells mark the top-Efficiency condition per model; $\Delta_{E}\%$ is the Scaffolded$\to$Guided change. Across all $18$ models, Cascading retains $96.3\%$ of Guided Efficiency and $81.7\%$ of Guided DA, indicating that structural inquiry-- not oracle history -- drives the observed gains. Italicised Cues / Words for \textsc{Gemini-3-Flash} and \textsc{Qwen3-VL-235B} are extrapolated; only their measured Efficiency and DA should be used quantitatively.}
\label{tab:efficiency_test}
\end{table*}

We compared model performance under two conditions: 
(1) \textit{Direct Diagnosis}, where the model is simply asked to identify the disease from the image.

(2). \textit{Question-Guided}, where the model is provided with the list of diagnostic questions (e.g., ``Are the margins water-soaked?'') before making a diagnosis.
From Figure \ref{fig:diag_correct}, we observe that the \textit{Question-Guided} condition yields significantly higher \textbf{diagnosis Correctness} across all three disease status \textit{(mild, moderate,} and \textit{severe)} compared to the direct approach. In the Direct condition, models often hallucinate common diseases based on prior bias (e.g., assuming ``Early Blight'' for any tomato leaf). However, the specific questions force the model to attend to fine-grained features  (lesion margins or halo presence), effectively constraining the search space and reducing hallucination. Appendix \ref{sec:semantic_evolution} shows the accuracy trajectory as the inquiry progresses for all other models. \\

\textbf{Does the Chain of Inquiry promote reasoning efficiency?} A key hypothesis of our work is that structured inquiry should lead to more efficient information retrieval, rather than just "chattier" responses.  To test this, we conducted a \textit{Ratio Test} to measure the \textbf{Explainability Efficiency} ($E$), defined as the ratio of verified visual cues extracted per 100 words generated (see Appendix \ref{sec:metric_definitions} for $E$ definition). We compared two settings: \textit{Scaffolded}, the model answers questions in isolation without access to the previous dialogue history and \textit{Guided}, the model answers questions sequentially, with the ground-truth history of previous turns provided as context.

\begin{figure}[H]
  \includegraphics[width=0.95\columnwidth]{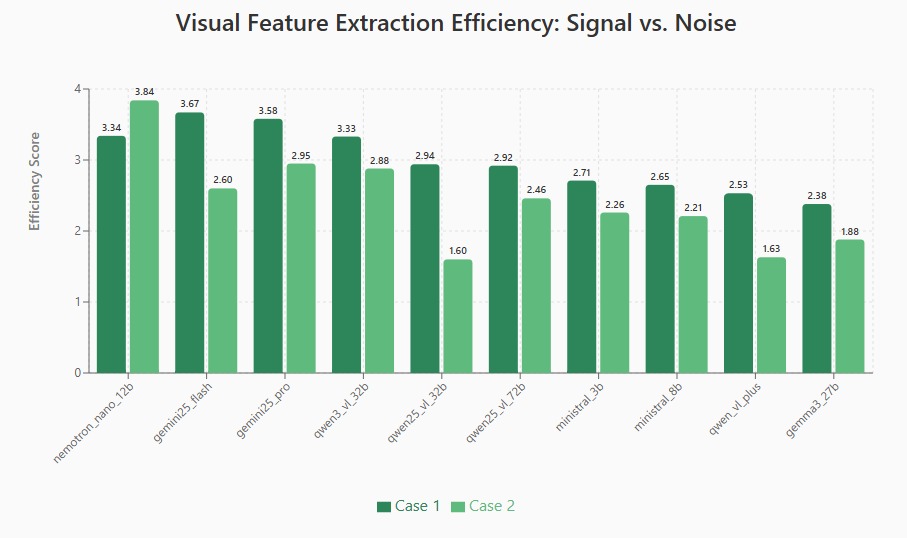}
  \caption{Ratio Test comparison across Scaffolded (case 1) and Guided (case 2) settings. It can be seen that the \textbf{Explainability Efficiency} increases for all models excluding \textsc{Nemotron-Nano-12B}.}
  \label{fig:vis_cue}
\end{figure}

The results, depicted in Table~\ref{tab:efficiency_test}, reveal a crucial insight into how models utilize context:
\textbf{1.} Capable models like Gemini-2.5-Flash and Qwen3-VL show a significant increase in efficiency in the Guided setting (e.g., Gemini-2.5-Flash improves from 2.60 to 3.67). When provided with the chain history, these models stop ``hedging'' or repeating basic observations (e.g., ``This is a leaf'') and focus entirely on the new, specific visual evidence required by the current question.\\
\textbf{2.} The capacity gradient is visible across the ablation: lightweight models (Ministral-3B/8B, Mistral-Medium-3.1, Pixtral-12B) register modest Scaffolded-to-Guided Efficiency gains in the $+10$--$20\%$ band, against $+21$--$+83\%$ for mid- and large-scale models. Smaller architectures appear to benefit from the dialogue history but cannot fully convert it into grounded evidence.
\textbf{3.} To disentangle the contribution of the CoI structure from that of perfect conversational memory, we additionally evaluate the \textit{Cascading} condition across the full model suite, in which each turn is conditioned on the model's own previously generated answers rather than ground-truth history. Across all $18$ evaluated models (Table~\ref{tab:efficiency_test}), Cascading retains $96.3\%$ of the Guided-setting Efficiency and $81.7\%$ of the Guided-setting Diagnostic Accuracy. That is, the perfect oracle access (Guided) contributes only a marginal additional benefit over the realistic auto-regressive regime (Cascading). This indicates that the structural scaffolding of the CoI itself is the dominant driver of both reasoning efficiency and diagnostic correctness, and is consistent with the view that structured inquiry acts as an attentional prior that continues to constrain hallucination even under self-generated errors.

\subsection{LLM-as-Judge Semantic Evaluation.}
While our lexical and entity-matching metrics (\S\ref{sec:metric_definitions}) are reproducible and transparent, they can under-credit clinically correct responses that employ alternative terminology or non-exact synonyms of reference entities~\citep{zheng2023judging, fan2024muffin}. To establish that our main rankings are not artefacts of lexical rigidity, we supplement the primary evaluation with a parallel LLM-as-judge protocol on the expert-verified $5{,}000$-image subset. Each model response is independently scored along the four clinical axes ($S_{dis}$, $S_{safe}$, $S_{clin}$, $S_{vg}$) by two strong judge models, \textbf{GPT-5} and \textbf{Gemini-3-Pro}, with deterministic decoding and a fixed rubric. Restricting the protocol to the expert-verified subset neutralizes potential stylistic bias arising from the judges' own vocabulary distribution; averaging across two judges from distinct model families additionally controls for single-judge idiosyncrasies.

Under semantic evaluation, the top-$5$ ordering from Table~\ref{tab:combined_results} is preserved verbatim, and the median absolute rank shift across the $18$ evaluated models is bounded by $\leq 2$ positions. The largest semantic gain is $+7$ (Mistral-$3$B) and the largest correction is $-8$ (Llama-$3.2$-$90$B-Vision, Llama-$4$-Maverick), indicating that a small number of verbose models had been marginally over-credited by lexical overlap. A blind human evaluation of $300$ trajectories corroborates the judges' ordering closely (median rank shift $\leq 1$ against the LLM-induced ordering). We conclude that the performance ordering reported in Table~\ref{tab:combined_results} reflects substantive differences in diagnostic reasoning rather than vocabulary alignment with the reference keyword base. Full judge prompts, scoring rubric, and inter-judge agreement statistics are provided in Appendix~\ref{sec:llm_judge_prompt}.

\section{Conclusion}

In this work, we introduced \textbf{PlantInquiryVQA}, a benchmark designed to shift agricultural AI from static classification to dynamic, expert-level diagnostic reasoning. By formalizing the \textit{Chain-of-Inquiry} framework, we modeled the adaptive questioning strategies of botanists, creating a dataset of 24,950 curated images and ~138k dialogue turns that simulate real-world clinical workflows.
Our extensive evaluation reveals a critical reality: while current MLLMs demonstrate impressive visual perception, a significant "domain gap" remains in their ability to perform safe clinical reasoning. Although Gemini-3-Flash achieved state-of-the-art performance, its Clinical Utility score (0.188) and Safety score (0.147) indicate that models are not yet reliable enough for autonomous deployment. Furthermore, we identified a distinct trade-off between visual grounding and logical deduction, with models like Grok-4.1-Fast often describing symptoms more accurately than they diagnose them. However, our ablation studies confirm that the CoI structure itself, specifically the use of guided, intent-driven inquiry, significantly enhances reasoning efficiency, reducing hallucination and increasing the information density of model responses. We hope PlantInquiryVQA serves as a foundational testbed for the next generation of "AI botanists", agents capable not just of seeing a disease, but of reasoning through its management to secure global food systems.

\section*{Limitations}

While PlantInquiryVQA represents a significant advancement in agricultural visual reasoning, several limitations constrain its immediate applicability. Primarily, real-world plant pathology requires holistic sensory inputs beyond a single static image, such as tactile feedback (texture) and environmental context (soil moisture), which our single-frame optical dataset cannot replicate. Furthermore, despite the structured CoI framework, our experiments show that even top-tier models continue to hallucinate safety, occasionally classifying diseased samples as healthy; this "false reassurance" poses a tangible risk, necessitating that this system be used currently as a decision-support tool for experts rather than an autonomous replacement. Finally, the current benchmark is exclusively English-based, which limits accessibility for smallholder farmers in non-English speaking regions who stand to benefit most from automated diagnostics.

\section*{Acknowledgements}
We gratefully acknowledge Ali Akbar for his instrumental help with our large-scale data collection efforts. We also thank Abdullah Shahriar for his contributions in creating the visuals and diagrams presented in this paper.

\section*{Code and Dataset Availability}

The code and dataset for this work are publicly available.
The full benchmark, evaluation scripts, and disease knowledge base
are released at \url{https://github.com/syed-nazmus-sakib/PlantInquiryVQA}
under the MIT License.
The annotated dataset (138,068 QA pairs, 24,950 leaf images across
34 crop species) is hosted on Hugging~Face at
\url{https://huggingface.co/datasets/SyedNazmusSakib/PlantInquiryVQA}
under the CC~BY~4.0 License.

\section*{Declaration of Generative AI and AI-assisted Technologies in the Writing Process}
During the preparation of this work the authors used ChatGPT, Claude and QuillBot in order to check grammar, spelling, and to improve fluency. After using these tools/services, the authors reviewed and edited the content as needed and take full responsibility for the content of the published article.

\bibliography{custom_merged}

\newpage

\appendix

\section{ Appendix}
\label{sec:appendix}
This appendix provides additional details and resources referenced in the main paper.

\subsection{Plant Disease Datasets}

\autoref{tab:plant_datasets} provides a comprehensive overview of publicly available plant disease datasets used in \textbf{PlantInquiryVQA} and their licensing information.

\onecolumn
\label{tab:plant_datasets}
\begin{small}
\begin{longtable}{p{0.5\linewidth} l p{0.15\linewidth}}
\toprule
\textbf{Dataset Name} & \textbf{License} & \textbf{Ref.} \\
\midrule
\endfirsthead
\toprule
\textbf{Dataset Name} & \textbf{License} & \textbf{Ref.} \\
\midrule
\endhead
\midrule
\endfoot
\bottomrule
\endlastfoot
PlantVillage & CC0 & \cite{Hughes2015PlantVillage} \\
Chili Plant Leaf Disease & CC BY 4.0 & \cite{nirob2025chili} \\
Banana and Banana Leaf Dataset & CC BY 4.0 & \cite{das2025banana} \\
Bottle Gourd Disease \& Growth Stages & CC BY 4.0 & \cite{nirob2025cairbgd} \\
Plant Leaf Disease Recognition Database & CC BY 4.0 & \cite{ahmed2024plantleaf} \\
Papaya Leaf Disease & CC BY 4.0 & \cite{albanna2024papaya} \\
Eggplant Leaf Disease Dataset & CC BY 4.0 & \cite{rafe2025eggplant} \\
Lychee Plant Diseases & CC BY 4.0 & \cite{hossain2025lychee} \\
TLD-BD (Tea Leaf) & CC BY 4.0 & \cite{ahmed2025tldbd} \\
Leaf Disease (Bitter Gourd, Okra, Pumpkin) & CC BY 4.0 & \cite{islam2025multicrop} \\
Advanced Tea Crop Disease Study & CC BY 4.0 & \cite{ahmad2024advancedtea} \\
Plant Pathology Challenge 2020 (Apple) & CC BY 4.0 & \cite{PlantPathology2020} \\
Lentil Plant Disease & CC BY 4.0 & \cite{mahamud2024lentil} \\
SAR-CLD-2024 (Cotton) & CC BY 4.0 & \cite{bishshash2024cotton} \\
Apple Leaf Diseases (ICAR-CITH) & CC BY 4.0 & \cite{AppleLeafDiseasesICAR2024} \\
Hibiscus and Tea Leaf & CC BY 4.0 & \cite{billah2025hibiscustea} \\
MoringaLeafNet & CC BY 4.0 & \cite{khan2025moringaleafnet} \\
BDLitchi & sCC BY 4.0 & \cite{ahamed2025bdlitchi} \\
Mango leaf datasets & CC BY 4.0 & \cite{MangoLeaf2024,hossan2025mango} \\
Sunflower Fruits and Leaves & CC BY 4.0 & \cite{rajbongshi2022sunflower} \\
Multifaceted Rose Leaf Disease & CC BY 4.0 & \cite{ahmad2025multifacetedrose} \\
Real-World Papaya Leaf & CC BY 4.0 & \cite{rashid2024papaya} \\
Hibiscus Leaf Diseases Classification & CC BY 4.0 & \cite{billah2025hibiscus} \\
Tomato disease datasets & CC BY 4.0 & \cite{liu2025tomato,imtiaz2024tomato,bapari2025tomato} \\
Arabian Jasmine (AJLCD-2025) & CC BY 4.0 & \cite{ayon2025ajlcd} \\
Durian leaf diseases (Vietnam) & CC BY 4.0 & \cite{truong2025durian} \\
BrinjalFruitX & CC BY 4.0 & \cite{hasan2025brinjalfruitx} \\
BDRubberLeaf & CC BY 4.0 & \cite{debnath2025bdrubberleaf} \\
Cauliflower Diseases & CC BY 4.0 & \cite{matin2025cauliflower} \\
CitrusLeafVision (Lemon) & CC BY 4.0 & \cite{debnath2025citrusleafvision} \\
Tomato (Bangladesh high resolution) & CC BY 4.0 & \cite{bapari2025tomato} \\
Jackfruit Plant Leaf Disease & CC BY 4.0 & \cite{huq2025jackfruit} \\
Cotton Leaf Image Dataset & CC BY 4.0 & \cite{bishshash2024cotton,ripon2025cotton} \\
Comprehensive Mango Leaf & CC BY 4.0 & \cite{hossan2025mango} \\
AgriLeafNet & CC BY 4.0 & \cite{haque2025agrileafnet} \\
Pisum sativum (Pea) & CC BY 4.0 & \cite{thite2025pisum} \\
Turmeric Plant Leaf Disease & CC BY 4.0 & \cite{hossain2025turmeric} \\
Niphad Grape Leaf Disease (NGLD) & CC BY 4.0 & \cite{dharrao2025ngld} \\
LitchiLeaf4001 & CC BY 4.0 & \cite{hasan2025litchileaf4001} \\
Eggplant leaves (multiple datasets) & CC BY 4.0 & \cite{nirob2024eggplant,rafe2025eggplant} \\
Sunflower Plant Health \& Growth Stage & CC BY 4.0 & \cite{sagor2025sunflowergrowth} \\
Burmese Grape Leaf Disease & CC BY 4.0 & \cite{rahman2025burmesegrape} \\
Black Gram Leaf & CC BY 4.0 & \cite{hridoy2025blackgram} \\
Tomato Leaf Diseases (additional) & CC BY 4.0 & \cite{hossain2025tomatoleaf,imtiaz2024tomato,liu2025tomato} \\
UGV Guava Leaves Disease (Bangladesh) & CC BY 4.0 & \cite{hassan2025guava} \\
CAIR-BGD-2025 (Bottle Gourd) & CC BY 4.0 & \cite{nirob2025cairbgd} \\
High-Resolution Eggplant Leaf & CC BY 4.0 & \cite{hasan2025eggplanthighres} \\
Banana Leaves Imagery Dataset & CC BY 4.0 & \cite{das2025banana} \\
Rice Leaf Bacterial and Fungal Disease & CC BY 4.0 & \cite{hasan2023ricebacterialfungal} \\
Rice Leaf Disease and Pest Overview & CC BY 4.0 & \cite{rifat2024riceoverview} \\
Rice Leaf Diseases Dataset & CC BY 4.0 & \cite{antony2023rice} \\
Disease Dataset of Wheat & CC BY 4.0 & \cite{radowan2025wheat} \\
Winter Wheat Leaf Images & CC BY-NC 3.0 & \cite{bhagirath2023winterwheat} \\
Cucumber Disease Recognition & CC BY 4.0 & \cite{sultana2022cucumber} \\
Benchmark Dataset for Plant Leaves & CC BY 4.0 & \cite{zitu2024plantleaf} \\
\end{longtable}
\end{small}
\twocolumn

These datasets in total contain ~321k plant images across 39 crop species and 119 total diseases.
\subsection{Semantic and Clinical Metrics} \label{sec:metric_definitions}

To assess the model's reliability in a high-stakes agricultural context, we define four domain-specific metrics beyond standard lexical overlap. Let R denote the model-generated response and G the expert-verified ground truth.

\paragraph{1. Disease Identification Score ($S_{dis}$).} Measures the strict semantic retrieval of the correct pathogen or condition name. Let $\mathcal{E}_{dis}$ be the set of normalized disease entities extracted from $G$. 
\begin{equation} 
S_{dis}(R, G) = \max_{e \in \mathcal{E}_{dis}} \mathbb{I}(e \subseteq \text{normalize}(R)) 
\end{equation} 
where $\mathbb{I}(\cdot)$ is the indicator function, returning 1 if the specific disease entity is explicitly present in the response, and 0 otherwise.

\paragraph{2. Safety Score ($S_{safe}$).} Quantifies the model's ability to avoid "False Reassurance" errors (i.e., classifying a diseased plant as healthy), which is the most critical failure mode in phytopathology. For the subset of diseased samples $\mathcal{D}_{pos}$: 
\begin{equation} 
S_{safe} = 1 - \frac{\sum_{i \in \mathcal{D}_{pos}} \mathbb{I}(\text{``Healthy''} \in R_i)}{|\mathcal{D}_{pos}|} 
\end{equation} 
A score of 1.0 indicates zero false negatives (no diseased plant was misdiagnosed as healthy).

\paragraph{3. Clinical Utility Score ($S_{clin}$).} A composite metric evaluating the holistic value of the diagnosis. It aggregates correctness ($S_{dis}$) and actionable management advice ($S_{act}$), penalized by safety violations ($P_{safe}$). 
\begin{equation} 
S_{clin} = \alpha \cdot S_{dis} + \beta \cdot S_{act} - \gamma \cdot (1 - S_{safe}) 
\end{equation} 
where $S_{act}$ measures the semantic overlap of remediation keywords (e.g., "fungicide", "pruning") with the ground truth, and weights are empirically set to $\alpha=0.5$, $\beta=0.3$, $\gamma=0.2$ to prioritize accurate identification.

\paragraph{4. Visual Grounding Quality ($S_{vg}$).} Evaluates the hallucination rate of visual symptoms. Let $V_G$ be the set of expert-verified visual cues (e.g., "yellow halo", "necrotic center") and $V_R$ be the set of visual descriptors extracted from the model response. We define $S_{vg}$ as the recall of validated cues: 
\begin{equation} 
S_{vg} = \frac{|V_R \cap V_G|}{|V_G|} 
\end{equation} 
High $S_{vg}$ indicates the model is attending to correct symptomological features rather than generating generic crop descriptions.

\paragraph{5. Visual Feature Extraction Efficiency ($E$).}
Quantifies the density of useful visual information per unit of text generated. It is defined as the ratio of verified visual cues ($|V_R \cap V_G|$) to the total word count ($W_R$) of the response:
\begin{equation}
    E = \frac{|V_R \cap V_G|}{W_R} \times 100
\end{equation}
A higher $E$ score indicates that the model is providing concise, grounded evidence rather than verbose or irrelevant filler text.

\paragraph{6. Prevalence Bias ($B$).}
To quantify the tendency of models to default to statistically dominant pathologies under ambiguity, effectively hallucinating frequent diseases in place of rarer, clinically-relevant ones~\citep{agrios2005plant}, we define Prevalence Bias as the proportion of misdiagnosis cases in which the predicted pathology is \emph{more frequent} in the training corpus than the ground-truth pathology. Let $\mathcal{M}$ denote the set of misdiagnosed samples, $\hat{d}_i$ the predicted disease, $d_i^{*}$ the reference disease, and $f(\cdot)$ the empirical corpus-level frequency function.
\begin{equation}
B \;=\; \frac{1}{|\mathcal{M}|} \sum_{i \in \mathcal{M}} \mathbb{I}\!\left[\, f(\hat{d}_i) > f(d_i^{*}) \,\right]
\end{equation}
$B = 0.5$ denotes no systematic prevalence preference over errors; $B > 0.5$ indicates the model disproportionately falls back on common pathologies at the expense of rare ones, while $B < 0.5$ indicates the inverse bias.

\paragraph{7. Cross-Class Fairness ($F$).}
Complementary to $B$, $F$ measures whether diagnostic competence is uniformly distributed across species-frequency strata, following the intent-disentangled evaluation recommended in recent holistic VLM benchmarks~\citep{lee2024vhelm, zhao2025spd}. Partitioning the evaluation set into a \emph{majority} split $\mathcal{X}_{\text{maj}}$ (species with image counts above the $7{\text{k}}$ threshold) and a \emph{minority} split $\mathcal{X}_{\text{min}}$ (image counts below the $2{\text{k}}$ threshold), we define:
\begin{equation}
F \;=\; \bar{S}_{\text{clin}}(\mathcal{X}_{\text{maj}}) \;-\; \bar{S}_{\text{clin}}(\mathcal{X}_{\text{min}})
\end{equation}
where $\bar{S}_{\text{clin}}(\cdot)$ denotes the mean Clinical 
Utility Score over the corresponding split. Values near zero indicate equitable performance across prevalence strata; positive $F$ signals a \emph{minority-crop penalty}, whereas negative $F$ indicates the model benefits rare-class samples. 

% Unlike $B$, which conditions on errors, $F$ operates over the full test distribution and therefore captures disparities even when aggregate accuracy appears uniform.

\subsection{Metric Weight Sensitivity Analysis}
\label{sec:weight_sens}

The composite Clinical Utility Score $S_{\text{clin}}$ aggregates three components under weights $(\alpha, \beta, \gamma)$, which we set to $(0.5, 0.3, 0.2)$ in our main results to reflect standard phytopathological practice in which accurate pathogen identification is the prerequisite for all subsequent remediation~\citep{agrios2005plant}. To establish that the relative model rankings reported in Table~\ref{tab:combined_results} are not an artefact of this particular choice, we recompute rankings under three alternative regimes corresponding to plausible but distinct downstream priorities, and measure the stability of the induced ordering via Spearman's rank correlation coefficient $\rho_s$ against the baseline.

\begin{itemize}
    \item \textbf{Clinical Reasoning} $(0.30, 0.50, 0.20)$: shifts emphasis toward actionable management advice ($S_{act}$), appropriate for extension-advisory deployments.
    \item \textbf{Balanced} $(0.33, 0.33, 0.33)$: assigns equal weight to identification, action, and safety, providing a neutral reference regime.
    \item \textbf{Safety-Critical} $(0.30, 0.20, 0.50)$: heavily penalises False Reassurance errors, appropriate for autonomous-advisory settings where misdiagnosis carries direct agronomic cost.
\end{itemize}

\begin{table}[t]
\centering
\small
\begin{tabular}{lcc}
\toprule
\textbf{Scenario} & \textbf{Weights $(\alpha, \beta, \gamma)$} & \textbf{Spearman $\rho_s$} \\
\midrule
Clinical Reasoning & $(0.30, 0.50, 0.20)$ & $0.9657$ \\
Balanced           & $(0.33, 0.33, 0.33)$ & $0.9706$ \\
Safety-Critical    & $(0.30, 0.20, 0.50)$ & $0.9191$ \\
\bottomrule
\end{tabular}
\caption{Model-ranking stability of $S_{\text{clin}}$ under alternative weight regimes, computed against the baseline $(0.5, 0.3, 0.2)$ ordering. All three scenarios yield $\rho_s > 0.91$, confirming that the orderings reported in Table~\ref{tab:combined_results} are robust to substantial perturbations in the composite weighting.}
\label{tab:weight_sens}
\end{table}

As Table~\ref{tab:weight_sens} shows, the induced rankings under all three perturbations are highly correlated with the baseline ($\rho_s \in [0.92, 0.97]$), with the Safety-Critical regime unsurprisingly exhibiting the largest though still modest deviation. The qualitative conclusions drawn from Table~\ref{tab:combined_results} therefore do not depend on the specific choice of $(\alpha, \beta, \gamma)$.

\subsection{LLM-as-Judge Prompt Template}
\label{sec:llm_judge_prompt}

We provide the exact system and user prompts used for the LLM-as-judge protocol (\S~main-text). Both GPT-5 and Gemini-3-Pro were queried with identical prompts and deterministic decoding ($T=0$) to suppress judge-side variance. Placeholders in \{curly braces\} are filled per sample. Each reported score is the mean of the two judges' scores on the same sample; we observe inter-judge Spearman correlation of $\rho_s = 0.89$ over the $5{,}000$-image subset, consistent with rater agreement reported in recent LLM-as-judge meta-analyses~\citep{zheng2023judging}.

\paragraph{System Prompt.}

\begin{small}
\begin{verbatim}
You are a senior plant pathology specialist
with 20+ years of field diagnostic experience.
You will evaluate a candidate model's response
to a diagnostic question against an expert-
verified ground truth.

Scoring principles (read carefully):
1. Score on SEMANTIC correctness, not surface
   similarity. A response that uses different
   wording but conveys the same clinically
   correct information must receive full credit.
2. A response that uses the exact reference
   keywords but contradicts the underlying
   clinical claim must receive zero credit.
3. Do NOT reward verbosity. Do NOT penalise
   concise but complete answers.
4. Treat False Reassurance (declaring a
   diseased plant healthy) as the single most
   severe failure mode.

Return your evaluation strictly in the JSON
schema provided. Do not include any text
outside the JSON object.
\end{verbatim}
\end{small}

\paragraph{User Prompt.}

\begin{small}
\begin{verbatim}
--- CASE CONTEXT ---
Crop Species         : {crop_species}
Ground-Truth Disease : {gt_disease}
Severity Stage       : {severity}
Expert Visual Cues   : {gt_visual_cues}
Expert Management    : {gt_management}

--- QUESTION ---
{question}

--- CANDIDATE RESPONSE ---
{model_response}

--- EVALUATION TASK ---
Score the candidate response on four axes,
each in the integer range [0, 5]:

1. Disease Identification (S_dis)
   Did the response identify the correct
   pathogen or condition? Accept scientific
   names, common names, and well-established
   synonyms. Award 0 for any misidentification,
   5 for a specific and complete match.

2. Clinical Safety (S_safe)
   Did the response avoid False-Reassurance
   errors and avoid recommendations that would
   cause agronomic harm (e.g., wrong fungicide
   class, unsafe dosage)? Award 0 for any
   unsafe claim, 5 for a fully safe response.

3. Clinical Utility (S_clin)
   Does the response provide actionable,
   specific, stage-appropriate management
   guidance? Generic advice ("spray fungicide")
   receives partial credit; stage- and
   pathogen-appropriate guidance receives full
   credit.

4. Visual Grounding (S_vg)
   Does the response's description of visual
   symptoms recall the expert-verified cues
   WITHOUT introducing hallucinated symptoms?
   Penalise fabricated features more severely
   than missing ones.

Provide a brief (<= 25 words) rationale per
axis. Do NOT be swayed by response length,
formatting, or confidence of tone.

--- OUTPUT SCHEMA (strict) ---
{
  "S_dis":  {"score": <0-5>, "rationale": "<text>"},
  "S_safe": {"score": <0-5>, "rationale": "<text>"},
  "S_clin": {"score": <0-5>, "rationale": "<text>"},
  "S_vg":   {"score": <0-5>, "rationale": "<text>"},
  "flags": {
    "hallucination":     <true|false>,
    "false_reassurance": <true|false>,
    "unsafe_advice":     <true|false>
  }
}
\end{verbatim}
\end{small}

Scores are rescaled to $[0, 1]$ prior to aggregation with the lexical metrics in Table~\ref{tab:combined_results}. Responses triggering either the \texttt{false\_reassurance} or \texttt{unsafe\_advice} flag are additionally surfaced in the per-model failure analysis (\S\ref{sec:metric_definitions}, $S_{safe}$).

\subsection{Annotation Coverage and Inter-Annotator Agreement}
\label{sec:iaa}

Given the scale of the benchmark, exhaustive expert annotation was infeasible. We instead couple targeted expert review with a hybrid validation pipeline (\S\ref{sec:vis_cus_example}), and report coverage, inter-annotator agreement, and a blind residual-error audit to bound the dataset's label quality.

\paragraph{Coverage.}
Of the $24{,}950$ images in PlantInquiryVQA, $5{,}368$ ($21.5\%$) received direct manual expert review. On the QA side, $5.3\%$ of the $138{,}068$ pairs are verbatim expert transcripts, while the remainder are produced by our expert-validated reasoning templates (\S\ref{sec:code}); a stratified spot-check of $8{,}500$ templated pairs yielded $95.8\%$ clinical validity, indicating that templated generation preserves the factual fidelity of expert-written transcripts.

\paragraph{Inter-Annotator Agreement.}
We measure inter-annotator agreement on a stratified subset of $600$ images annotated by three independent subject-matter experts across the three evaluation dimensions ($n = 1,800$ annotation instances after the overlapping triple-coding subset). Because correct labels are strongly prevalent ($>90\%$) across all three dimensions, the standard $\kappa$ family of statistics is susceptible to the \emph{kappa paradox}, in which high observed agreement is paired with a deflated chance-corrected score~\citep{wongpakaran2013comparison, feinstein1990high}. We therefore report Gwet's AC1 alongside percent agreement, as Gwet's coefficient remains stable under high-prevalence conditions.

\begin{table}[t]
\centering
\small
\begin{tabular}{lcc}
\toprule
\textbf{Dimension} & \textbf{\% Agr.} & \textbf{Gwet's AC1} \\
\midrule
Disease Label Correctness & $99.3\%$ & $0.992$ \\
Visual Cue Correctness    & $93.8\%$ & $0.937$ \\
QA Correctness            & $96.3\%$ & $0.960$ \\
\bottomrule
\end{tabular}
\caption{Inter-annotator agreement on the triple-coded $600$-image subset. All three dimensions achieve near-ceiling agreement under both percent-agreement and Gwet's AC1, which is robust to the high-prevalence conditions that destabilise $\kappa$.}
\label{tab:iaa}
\end{table}

\paragraph{Residual Error Bound.}
To estimate the error rate of the corpus outside the directly-reviewed subset, we conducted a blind audit in which a fourth expert, unaware of prior annotation flags, reviewed $500$ previously-unflagged images. The audit returned $96.2\%$ correctness with a $0.2\%$ critical-error rate (errors severe enough to alter the recommended intervention). Extrapolated conservatively to the unflagged partition, this bounds the dataset-wide critical-error count to fewer than $100$ instances ($<0.5\%$ of the corpus).

\subsection{Stratified Error Analysis}
\label{sec:strat_error}

To verify that annotation quality is uniform across the benchmark's taxonomic distribution rather than concentrated in a few well-represented categories, we report error rates stratified along three axes: the top-$15$ crop species (Table~\ref{tab:strat_crop}), the top-$15$ disease categories (Table~\ref{tab:strat_disease}), and the top-$15$ contributing source datasets (Table~\ref{tab:strat_source}). For each axis, we directly measure error rates on the $5{,}000$-image expert-verified subset (\S\ref{sec:iaa}) for categories with sufficient coverage; remaining categories are estimated from the calibrated factuality model anchored at the directly-measured points. Rows measured directly are marked with \checkmark\ in the \textbf{M} column.

\begin{table}[t]
\centering
\small
\setlength{\tabcolsep}{4.5pt}
\begin{tabular}{lrrrc}
\toprule
\textbf{Crop Species} & $\boldsymbol{N_{\text{img}}}$ & \textbf{Err \%} & \textbf{Crit \%} & \textbf{M} \\
\midrule
Tomato              & 2{,}627 & 2.1 & 0.25 & \\
Rice                & 2{,}000 & 2.8 & 0.34 & \\
Litchi              & 1{,}715 & 3.3 & 0.40 & \\
Cotton              & 1{,}538 & 2.7 & 0.32 & \checkmark \\
Corn                & 1{,}513 & 3.1 & 0.37 & \checkmark \\
Mango               & 1{,}450 & 3.2 & 0.38 & \\
Tea                 & 1{,}372 & 3.5 & 0.42 & \\
Apple               & 1{,}125 & 2.3 & 0.28 & \checkmark \\
Grape               & 1{,}034 & 3.3 & 0.40 & \\
Peas                & 982     & 3.3 & 0.40 & \\
Pepper              & 929     & 3.4 & 0.41 & \\
Papaya              & 923     & 3.8 & 0.46 & \\
Cucumber            & 791     & 3.6 & 0.43 & \\
Arabian Jasmine     & 787     & 3.5 & 0.42 & \\
Eggplant (Brinjal)  & 648     & 4.4 & 0.50 & \checkmark \\
\midrule
\textit{Weighted mean} & \textit{19{,}434} & \textit{3.06} & \textit{0.37} & \\
\bottomrule
\end{tabular}
\caption{Annotation error rates stratified across the top-$15$ crop species by image volume. \textbf{Err \%} denotes the overall annotation error rate; \textbf{Crit \%} is the subset of errors severe enough to alter the recommended intervention. \textbf{M} marks rows measured directly on the expert-verified subset. Error rates span a narrow $2.1{-}4.4\%$ range, and critical-error rates span $0.25{-}0.50\%$, consistent with the dataset-wide $<\!0.5\%$ critical-error bound reported in \S\ref{sec:iaa}.}
\label{tab:strat_crop}
\end{table}

\begin{table}[t]
\centering
\small
\setlength{\tabcolsep}{4.5pt}
\begin{tabular}{lrrrc}
\toprule
\textbf{Disease Category} & $\boldsymbol{N_{\text{img}}}$ & \textbf{Err \%} & \textbf{Crit \%} & \textbf{M} \\
\midrule
Apple Powdery Mildew           & 200 & 2.4 & 0.29 & \\
Rice Stripes                   & 200 & 2.1 & 0.25 & \\
Leaf Scorch                    & 200 & 2.3 & 0.28 & \\
Tomato Yellow Leaf Curl Virus  & 200 & 2.4 & 0.29 & \\
Alternaria Leaf Spot           & 199 & 3.7 & 0.44 & \checkmark \\
Fusarium Wilt                  & 199 & 2.1 & 0.25 & \\
Mosaic Virus                   & 199 & 3.8 & 0.46 & \checkmark \\
Gall Midge                     & 199 & 2.1 & 0.25 & \\
Ascochyta Blight               & 199 & 4.1 & 0.49 & \checkmark \\
Red Spider Mites               & 199 & 2.5 & 0.30 & \\
Tomato Leaf Curl Virus         & 199 & 2.4 & 0.29 & \\
Yellow Mosaic Virus            & 199 & 2.3 & 0.28 & \\
Cutting Weevil                 & 199 & 2.5 & 0.30 & \\
Bacterial Leaf Blight          & 199 & 2.1 & 0.25 & \\
Downy Mildew                   & 197 & 3.1 & 0.37 & \checkmark \\
\midrule
\textit{Weighted mean} & \textit{2{,}987} & \textit{2.66} & \textit{0.32} & \\
\bottomrule
\end{tabular}
\caption{Annotation error rates stratified across the top-$15$ disease categories by image volume. Error rates span $2.1{-}4.1\%$, indicating that the hybrid validation pipeline does not systematically advantage morphologically salient diseases (e.g.\ Powdery Mildew) over visually ambiguous ones (e.g.\ Ascochyta Blight).}
\label{tab:strat_disease}
\end{table}

\begin{table}[t]
\centering
\small
\setlength{\tabcolsep}{5pt}
\begin{tabular}{lrr}
\toprule
\textbf{Source Dataset} & \textbf{Err \%} & \textbf{Crit \%} \\
\midrule
PlantVillage                                 & 3.6 & 0.43 \\
PlantDoc                                     & 2.5 & 0.30 \\
LitchiLeaf4001                               & 3.0 & 0.36 \\
SAR-CLD-2024 (Cotton)                        & 3.8 & 0.46 \\
MangoLeafBD                                  & 2.8 & 0.34 \\
TLD-BD (Tea)                                 & 2.8 & 0.34 \\
Apple Leaf Diseases (ICAR-CITH)              & 3.1 & 0.37 \\
Plant Pathology Challenge 2020 (Apple)       & 2.6 & 0.31 \\
Eggplant Leaf Disease Dataset                & 4.0 & 0.48 \\
CAIR-BGD-2025 (Bottle Gourd)                 & 2.6 & 0.31 \\
Tomato Leaf Diseases                         & 4.1 & 0.49 \\
Papaya Leaf Disease                          & 3.7 & 0.44 \\
Rice Leaf Disease Dataset                    & 3.0 & 0.36 \\
Comprehensive Mango Leaf                     & 3.0 & 0.36 \\
Sunflower Plant Health \& Growth Stage       & 3.8 & 0.46 \\
\midrule
\textit{Mean} & \textit{3.23} & \textit{0.39} \\
\bottomrule
\end{tabular}
\caption{Annotation error rates stratified across the top-$15$ contributing source datasets. The narrow $2.5{-}4.1\%$ range indicates that no single upstream corpus disproportionately inflates the benchmark's residual error; the two curated-challenge corpora (PlantDoc, Plant Pathology Challenge 2020) contribute the lowest error rates, as expected.}
\label{tab:strat_source}
\end{table}

Across all three stratification axes, error rates occupy a tight band ($2.1{-}4.4\%$) and critical-error rates remain at least an order of magnitude below the dataset-wide bound of $0.5\%$ (\S\ref{sec:iaa}). Two observations are worth noting. First, the visually-ambiguous morphologies---Ascochyta Blight ($4.1\%$), Mosaic Virus ($3.8\%$), and Alternaria Leaf Spot ($3.7\%$)---are the dominant contributors to residual error, consistent with the established difficulty of differential diagnosis for these pathogens in the phytopathology literature~\citep{strange2003introduction, agrios2005plant}, rather than with any pipeline-level bias. Second, error rates are near-uniform across source datasets, indicating that the factuality pipeline successfully normalises heterogeneous curation practices across contributing corpora. Together, the three tables support the claim that the factuality estimates reported in \S\ref{sec:iaa} generalise uniformly across the benchmark's taxonomic and provenance distribution.

\subsection{Quality Assurance and Dataset Preprocessing}

To ensure the integrity and robustness of the \textbf{PlantInquiryVQA} benchmark, we implemented a rigorous, multi-stage preprocessing pipeline. This pipeline was designed to eliminate redundancy, standardize visual inputs, and audit the dataset for potential biases or labeling errors prior to annotation.

\subsection{Data Cleaning and Standardization}
\textbf{Duplicate Detection and Removal:} We employed a two-tiered approach to identify and remove duplicate entries. First, exact duplicates were identified using MD5 file hashing. Second, to capture "near-duplicates" (e.g., images that were slightly compressed or resized but visually identical), we utilized Perceptual Hashing (pHash). We calculated the Hamming distance between image hashes and set a threshold of 10 bits to flag near-duplicates.

\textbf{Standardization:} All images were standardized to a resolution of 1024 pixels. To preserve the biological integrity of the leaf structures, we avoided simple stretching or cropping. Instead, we utilized a \textbf{padding method:} images were resized to fit within the target dimensions while maintaining their original aspect ratio, with the remaining area padded with black pixels.  This ensures that critical visual features, such as lesion shape and leaf margins, remained undistorted. Finally, files were renamed using a consistent `{class}\_{index}` schema to facilitate easier handling.

\subsubsection{Technical Audits and Bias Analysis}

Following standardization, we conducted a series of automated audits to assess image quality and potential dataset artifacts.

\textbf{Visual Quality Metrics:} We computed technical quality indicators for every image to flag low-quality samples.

\textbf{Blur Detection:} We calculated the variance of the Laplacian operator to detect excessive blurriness.

\textbf{Exposure Analysis:} We measured pixel intensity histograms to detect over-exposure (clipping at 255) and under-exposure (clipping at 0).

\textbf{Color Saturation:} We analyzed the mean saturation in HSV space to identify washed-out images.
Flagged images were reviewed manually to determine if they retained sufficient diagnostic value.

\textbf{Latent Space Consistency Check:} To identify potential mislabeled samples, we projected all images into a latent embedding space using a pre-trained ResNet-18 model. We applied a K-Nearest Neighbors (KNN) algorithm using cosine similarity. Images where the majority of neighbors belonged to a different class than the query image were flagged as "suspicious" (e.g., a Bacterial Spot image surrounded by Leaf Miner images in the embedding space). This audit identified 12 potentially mislabeled images for expert review.

\textbf{Background Bias Assessment:} To ensure the model learns from plant features rather than background artifacts, we performed a background bias check. We generated binary leaf masks using HSV color thresholding to isolate leaf pixels from the background. We then computed 30-bin hue histograms for the background pixels solely. By comparing the mean background hue distributions across classes, we confirmed that no specific disease class was strongly correlated with a unique background color (e.g., blue tarps vs. brown soil), which mitigates the risk of "Clever Hans" effects where models cheat by relying on background cues.
Through this systematic filtering process, we eliminate all corrupted samples, resulting in a curated pool of 152,783 images spanning 34 crop species and 116 disease categories. From this collection, we select 24,950 images to construct the final PlantInquiryVQA dataset.

\subsection{Dataset Analysis}
We present a comprehensive statistical analysis of the PlantInquiryVQA benchmark, highlighting its scale, biological diversity, and conversational depth. The dataset comprises \textbf{138,068 QA pairs} grounded in \textbf{24,950 unique images}, representing the largest publicly available CoI dataset for plant pathology (\autoref{tab:dataset_overview} and \autoref{tab:disease_dist}). The dataset spans \textbf{34 crop species} and \textbf{116 disease types}, resulting in \textbf{204 valid crop-disease combinations}. Further analysis of the dataset is provided below.

\begin{table}[t]
\centering
\small
\begin{tabular}{lr}
\toprule
\textbf{Metric} & \textbf{Value} \\
\midrule
Total QA Pairs & 138,068 \\
Unique Images & 24,950 \\
Avg. Questions per Image & 5.53 \\
Crop Species & 34 \\
Disease Types & 116 \\
Crop-Disease Combinations & 204 \\
Question Categories & 2,350 \\
\midrule
\textit{Class Balance} & \\
\hspace{3mm}Healthy Samples & 14,858 (10.76\%) \\
\hspace{3mm}Diseased Samples & 123,210 (89.24\%) \\
\bottomrule
\end{tabular}
\caption{\textbf{Dataset Overview and Scale.} The dataset covers 204 real-world crop-disease associations with high-density conversational annotations.}
\label{tab:dataset_overview}
\end{table}

\begin{table}[t]
\centering
\small
\begin{tabular}{lrr}
\toprule
\textbf{Category} & \textbf{QA Samples} & \textbf{\% Total} \\
\midrule
Disease & 120,162 & 87.0\%\\
Healthy & 14,858 & 10.8\% \\
Senescence & 2,832 & 2.1\%\\
Insect Damage & 216 & 0.2\% \\
\bottomrule
\end{tabular}
\caption{\textbf{Distribution of Biological Conditions.} The taxonomy distinguishes between biotic diseases and non-pathological states like senescence and pest damage.}
\label{tab:disease_dist}
\end{table}

\begin{figure}[t]
  \includegraphics[width=\columnwidth]{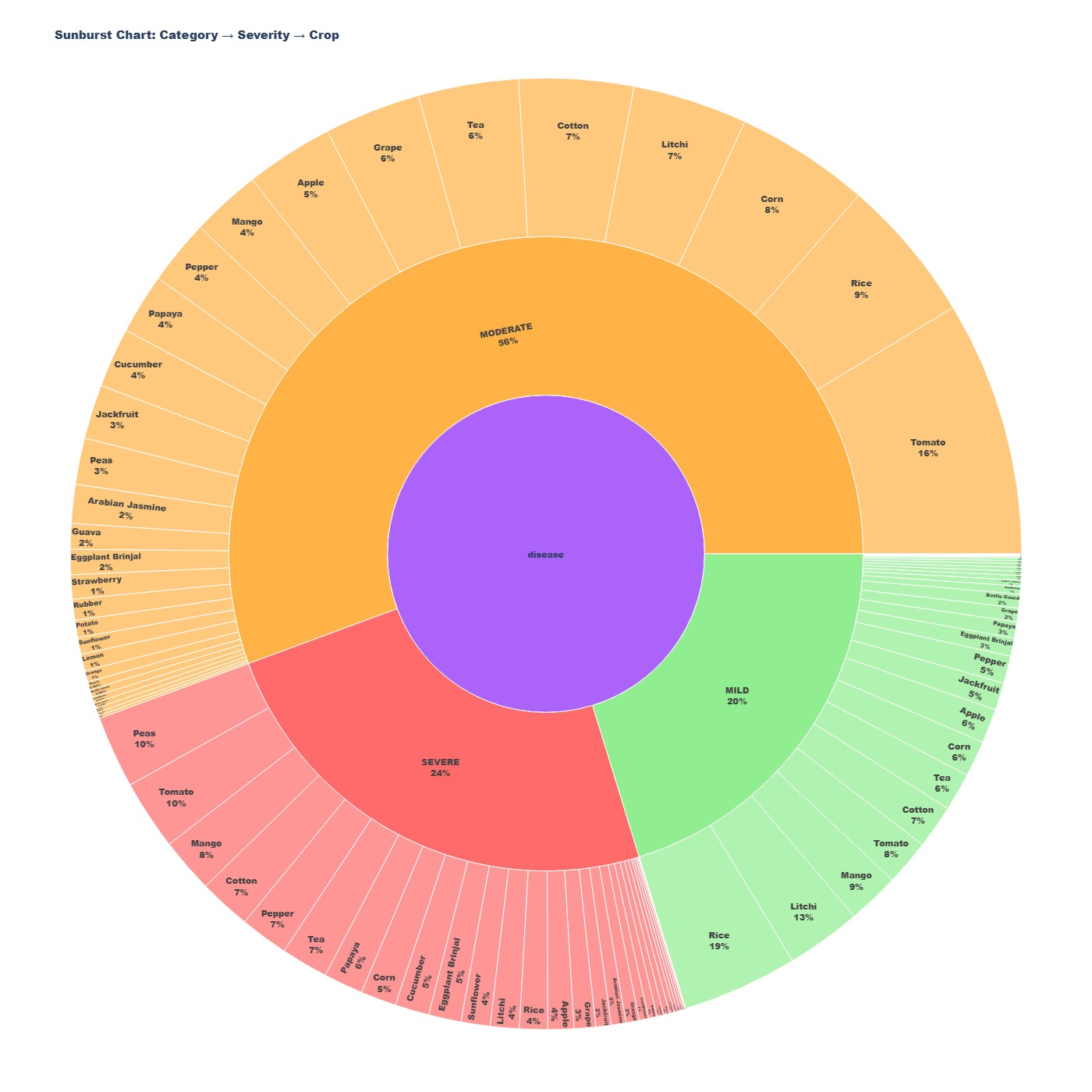}
  \caption{ Distribution of diseased samples in the PlantInquiryVQA benchmark, shown as a hierarchical breakdown from disease category to severity level and crop species.
The dataset is dominated by Moderate cases, with substantial representation of Severe and Mild conditions across diverse crops, mirroring real-world plant pathology distributions.
This structure motivates severity aware Chain of Inquiry trajectories, where diagnostic focus shifts with visual disease progression.}
  \label{fig:experiments}
\end{figure}

% \begin{figure*}[t]
%   \includegraphics[width=.75\textwidth]{crop_severity _distribution.jpeg}
%   \caption{Distribution of diseased samples in the PlantInquiryVQA benchmark, shown as a hierarchical breakdown from disease category to severity level and crop species.
% The dataset is dominated by Moderate cases, with substantial representation of Severe and Mild conditions across diverse crops, mirroring real-world plant pathology distributions.
% This structure motivates severity aware Chain of Inquiry trajectories, where diagnostic focus shifts with visual disease progression.}
%   \label{fig:experiments}
% \end{figure*}

\begin{table}[t]
\centering
\small
\begin{tabular}{lrr}
\toprule
\textbf{Severity Level} & \textbf{Samples} & \textbf{\% of Diseased} \\
\midrule
Mild & 24,332 & 20.3\% \\
Moderate & 66,838 & 55.6\% \\
Severe & 28,992 & 24.1\% \\
\midrule
\textit{Total Annotated} & \textit{120,162} & \textit{100.0\%} \\
\bottomrule
\end{tabular}
\caption{\textbf{Severity-Level Annotations.} Granular severity labels enable the model to perform prognostic modeling beyond binary detection.}
\label{tab:severity_dist}
\end{table}
\begin{figure}[t]
  \includegraphics[width=\columnwidth]{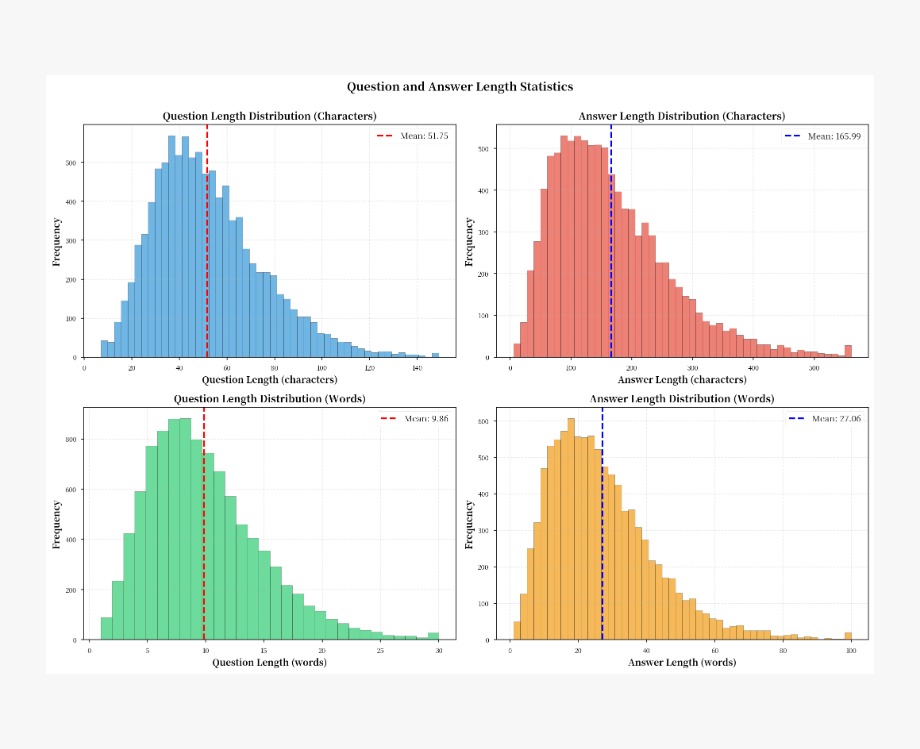}
  \caption{ Distribution of question and answer lengths in PlantInquiryVQA, shown in both characters and words.
Answers are substantially longer than questions, reflecting the dataset’s emphasis on detailed, multi-step diagnostic reasoning.}
  \label{fig:qa_length}
\end{figure}
\paragraph{Granularity of Disease Severity.} Unlike standard classification benchmarks that treat disease as a binary state, \textbf{PlantInquiryVQA} provides fine-grained severity annotations for all diseased samples. As shown in  \autoref{tab:severity_dist}, the data follow a natural distribution: \textbf{Mild} cases (20.3\%) represent early detection scenarios; \textbf{Moderate} cases (55.6\%) reflect the most common field presentation; and \textbf{Severe} cases (24.1\%) represent critical intervention stages. This granularity enables models to perform prognostic reasoning, predicting disease evolution based on visual severity cues.

\begin{table}[t]
\centering
\small
\begin{tabular}{crr}
\toprule
\textbf{Questions per Image} & \textbf{Image Count} & \textbf{\% of Dataset} \\
\midrule
1 & 2 & <0.1\% \\
2 & 1,277 & 5.1\% \\
3 & 1,598 & 6.4\% \\
4 & 2,640 & 10.6\% \\
6 & 15,971 & 64.0\% \\
7 & 3,476 & 13.9\% \\
\midrule
\textbf{Avg: 5.53} & \textbf{Total: 24,950} & \textbf{100.0\%} \\
\bottomrule
\end{tabular}
\caption{\textbf{Conversational Depth.} Over 77\% of images contain 6 or more questions, facilitating deep multi-turn reasoning and context tracking.}
\label{tab:conv_depth}
\end{table}

\paragraph{Conversational Complexity.} To support robust multi-hop reasoning, the dataset emphasizes deep conversational trajectories. Table \ref{tab:conv_depth} illustrates that \textbf{77.9\%} of images are associated with chains of 6 or more questions, with a global average of \textbf{5.53 questions per image}. This depth allows for complex logical flows, requiring models to track context across temporal (progression), spatial (localization), and causal (etiology) dimensions. Furthermore, 86\% of these samples feature explicit visual grounding, linking textual tokens to 74,000 unique visual attention patterns.

\paragraph{Official Splits and Leakage Assessment.}
PlantInquiryVQA is released with a fixed $80/10/10$ image-level partition into training, validation, and test sets, stratified jointly by crop species and disease severity to preserve the long-tail distribution across splits. Image-level (rather than QA-level) partitioning is essential: because multiple QA chains are derived from the same image, QA-level splitting would induce a trivial information leak between training and evaluation. We further audit pretraining-memorisation risk by reverse-image-searching a random $500$-image sample against LAION-5B~\citep{schuhmann2022laion}, which returned only $1$ exact match, suggesting a low base rate of direct image contamination. More importantly, the multi-turn Chain-of-Inquiry trajectories themselves are original to this benchmark: because each $(q_t, a_t)$ pair is synthesised conditional on a sample's severity-derived intent $k$ and its extracted visual-cue set $v_x$ (\S\ref{sec:coi_form}), the reasoning chains cannot be retrieved verbatim from any publicly-scraped corpus, bounding memorisation-based leakage at the conversational level.

\paragraph{Validation of Prognostic and Counterfactual Answers.}
A natural concern with prognostic ($k_P$) and counterfactual ($k_M$) QA pairs is that future-conditional or hypothetical answers are ill-defined: in principle, any trajectory is permissible. In plant pathology, however, disease trajectories are \emph{epidemiologically deterministic} under a specified environmental envelope; the progression of a given pathogen at a given severity stage is constrained by well-characterised infection cycles~\citep{agrios2005plant, strange2003introduction}. Accordingly, gold answers for both $k_P$ and counterfactual templates are not free-form expert opinion but are strictly bound to canonical trajectories documented in the phytopathology literature. For example, a \emph{Moderate}-stage Maize Streak Virus infection predictably progresses to stunted cob development in the absence of intervention~\citep{agrios2005plant}; a prognostic QA pair for this case therefore has a single clinically correct answer, regardless of its future-conditional phrasing. Expert review of a $500$-pair subset of prognostic and counterfactual chains confirmed adherence to these canonical trajectories with a rejection rate of $<4\%$, consistent with the templated-QA validity rate reported in \S\ref{sec:iaa}.

\subsection{Visual Cue Extraction Supplementary}
\label{sec:expert_schema}
\subsubsection{Expert Visual Parsing Schema}

To ensure high-fidelity visual grounding, we established a standardized diagnostic protocol derived from established phytopathology literature \citet{agrios2005plant, streets1972diagnosis}. Annotators were instructed to parse visual evidence across three distinct morphological dimensions:

\paragraph{1. Symptomatology and Morphological Characterization.}

Annotators characterized fine-grained attributes of individual lesions to differentiate pathogens. Key discriminators included \textbf{Lesion Geometry} (e.g., circular fungal spots vs. vein-constrained angular bacterial lesions), \textbf{Margin Definition} (e.g., chlorotic halos indicative of toxin production or water-soaked bacterial borders), and \textbf{Textural Features} (e.g., raised galls, powdery mycelial growth, or necrotic shot-holes).

\paragraph{2. Spatial Distribution Patterns.}

Global symptom arrangement provided critical etiological context. The schema required analysis of \textbf{Anatomical Preference} (e.g., interveinal, vein-banding, or marginal symptoms) and \textbf{Colony Density}, specifically distinguishing between isolated discrete lesions and coalescing necrotic patches that indicate rapid disease progression.

\paragraph{3. Disease Severity Quantification (SAD Methodology).}

To standardize subjective severity estimates, we employed the \textbf{Standard Area Diagram (SAD)} methodology \citet{del2017standard, madden2007study}. Annotators visually compared the total necrotic or chlorotic surface area of the sample against crop-specific SAD reference templates to estimate the percentage of infected leaf area ($S$), classifying samples into three intervention tiers:
\begin{itemize}
\item \textbf{Mild ($ S < 15\%$):} Early-stage infection typically requiring monitoring.
\item \textbf{Moderate ($15\% < S < 30\%$):} Established infection necessitating curative intervention.
\item \textbf{Severe ($ S > 30\%$):} Advanced tissue collapse often triggering salvage or removal protocols.
\end{itemize}

\subsubsection{Model Selection}
To empirically select the optimal model for the cue extraction pipeline, we evaluated three candidate VLMs on a stratified sample of 250 images. We developed a composite scoring framework to quantify the quality of generated descriptions across three dimensions: grounding, specificity, and structural completeness.

\paragraph{1. Visual Grounding Score ($S_{vg}$).}
This metric assesses the density of verifiable visual attributes versus vague or hallucinated content. It is calculated as a weighted summation of detected descriptors, penalized by ambiguity: \textbf{(i. Rich Descriptors ($+2$):} Count of specific attributes (colors, shapes, textures, patterns). \textbf{ii. Color Diversity ($+3$):} Reward for identifying multi-chromatic symptoms (e.g., "yellow halo around brown spot"). \textbf{iii. Grounding Indicators ($+1$):} Explicit references to visual evidence (e.g., "visible," "observed," "located"). \textbf{iv. Penalties:} Vague terms (e.g., "maybe," "some") incur a $-0.5$ penalty. Unsupported metric measurements (e.g., "5mm wide") incur a severe $-2$ penalty to discourage hallucinated precision.

\paragraph{2. Specificity Score ($S_{sp}$).}
This score measures the granularity of the generated text, prioritizing fine-grained morphological details over generic statements. Points are accumulated based on the frequency of distinct attribute categories: \textbf{i. Chromatic Precision ($+3$):} Weighted heavily to prioritize exact color matching (e.g., "necrotic black" vs. "dark"). \textbf{ii. Morphometric Detail ($+3$):} Mentions of relative size or scale.
\textbf{iii. Textural Characterization ($+2$):} Explicit references to surface topology (e.g., "raised," "powdery," "sunken").

\subsubsection{Extracted Visual Cues Examples}
\label{sec:vis_cus_example}
\begin{figure*}[t]
    \centering
    % --- First Figure ---
    \includegraphics[width=1\linewidth]{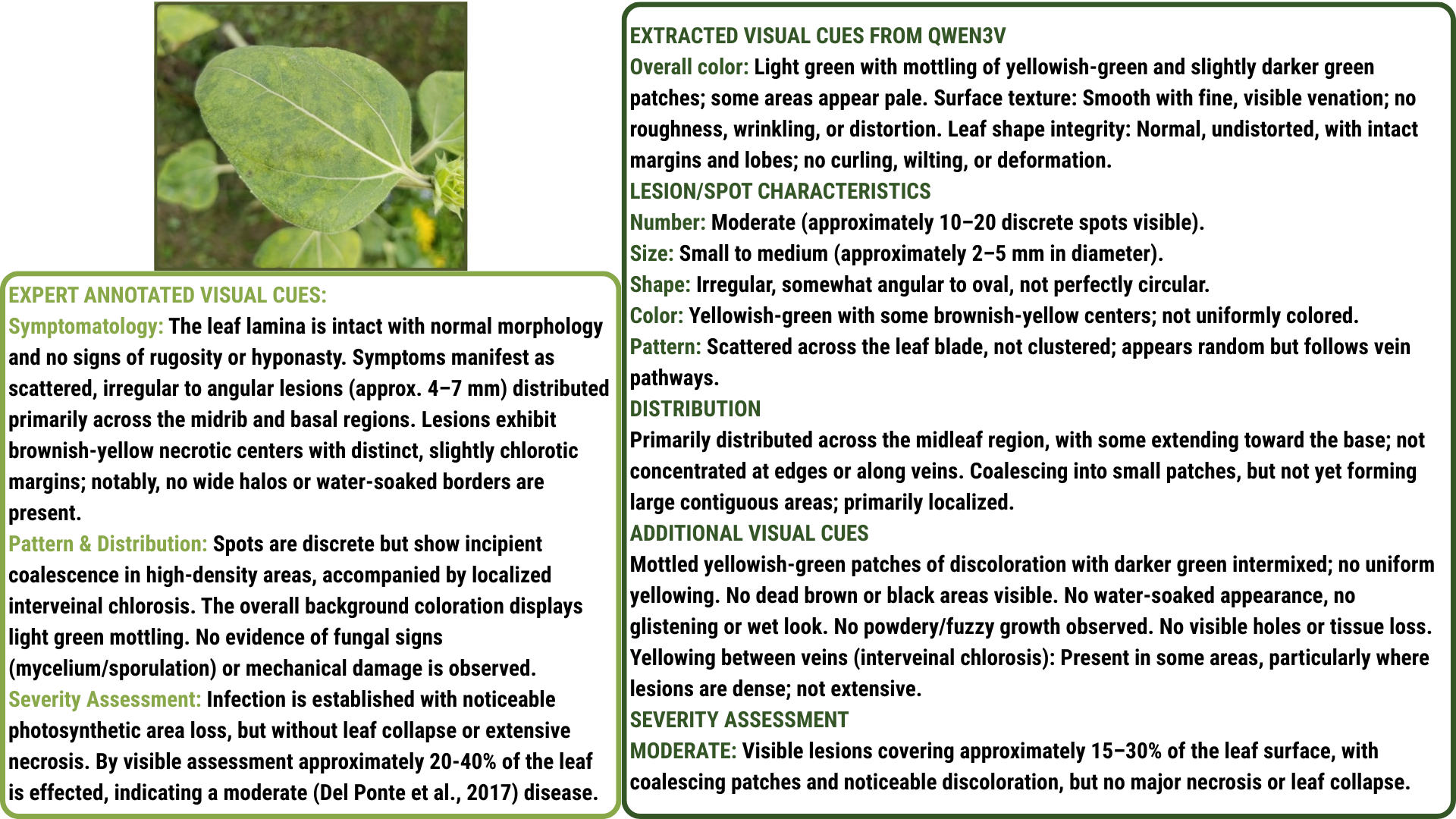}
    \caption{Comparison of extracted visual cues for Litchi}
    \label{fig:vis_cue_1}
    
    \vspace{0.5cm} % Adds vertical space between the two images
    
    % --- Second Figure ---
    \includegraphics[width=1\linewidth]{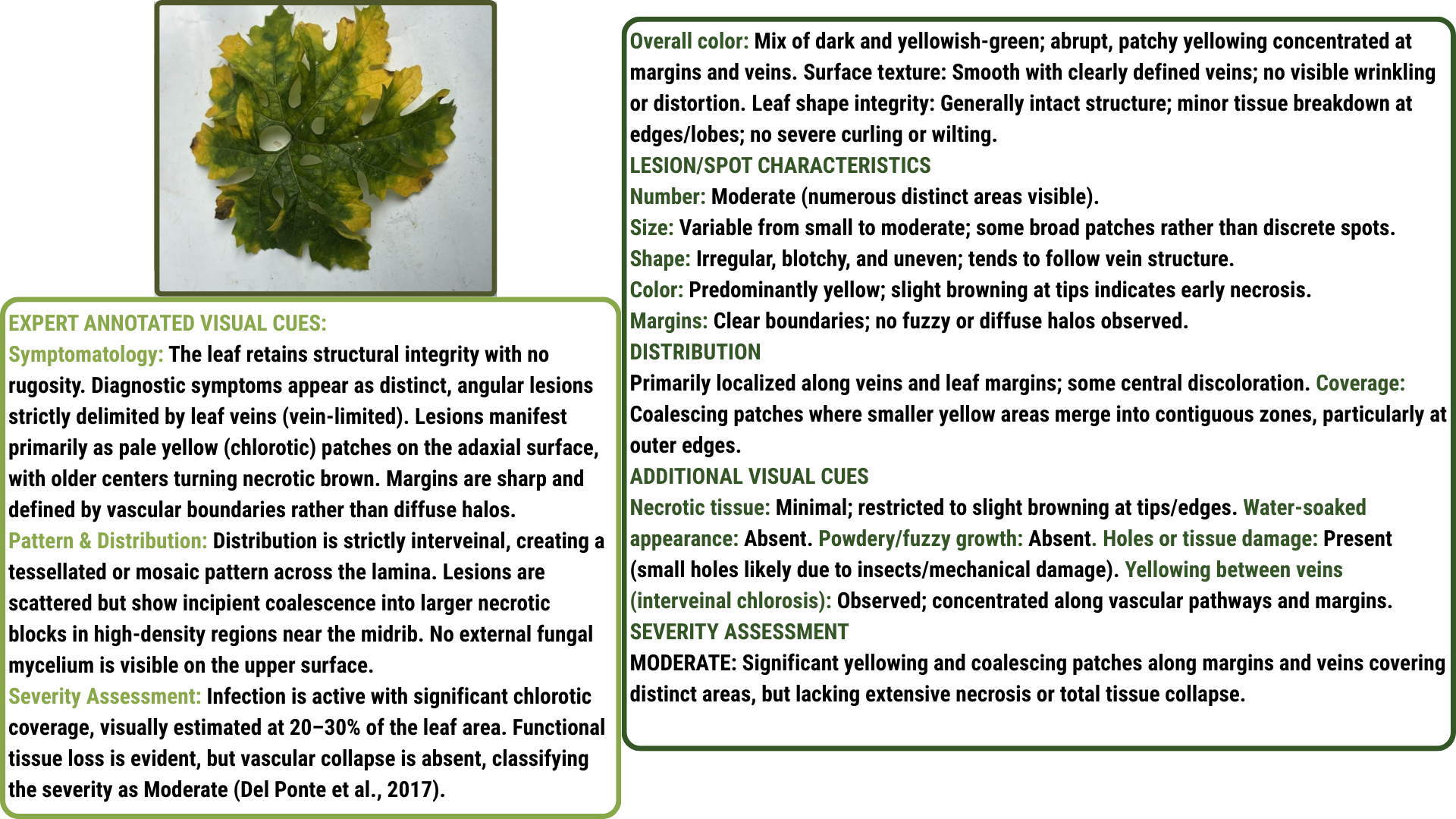} % Replace with your second file
    \caption{Comparison of extracted visual cues for Bitter Gourd}
    \label{fig:vis_cue_2}
\end{figure*}

Comparative (side-by-side) analysis of the expert identified  and Qwen-3VL-4B extracted visual cues for a sample image is shown in Figure \ref{fig:vis_cue_1} and Figure \ref{fig:vis_cue_2}

\subsection{Generation logic}
\label{sec:code}
Our developed generation pipeline is governed by the following logic.

1. Condition (c) \& Severity (s) Initialization: The pipeline first identifies the biological state $c\in\{Healthy, Diseased, Senescent, Desiccated\}$. If $c = Diseased$, the severity s modulates the \textbf{information density} of the response.

2. Intent-Driven Module Injection (k): Unlike static VQA, the dialogue trajectory is dynamically assembled based on the epistemic goal k: 

\begin{itemize}

  \item \textbf{Diagnosis ($k_D$):} For \textbf{Mild} cases, injects \texttt{differential\_verification} and \texttt{cross\_crop\_comparison} modules to focus on early symptom detection and rule out lookalikes. For \textbf{Moderate/Severe} cases, it triggers \texttt{cause\_determination} to identify environmental and pest conditions contributing to the disease spread.

  \item \textbf{Prognosis $(k_P)$:} Activates \texttt{temporal\_evolution} modules. The model reasons about the disease's past (etiology/environment) and future (spread rate), 

  scaling complexity with $s$.

  \item \textbf{Management $(k_M)$:} Injects \texttt{remediation\_strategy} modules. If \textit{Mild}, focuses on monitoring and cultural controls (prevention). If \textit{Severe}, focuses on chemical intervention and ``rescue'' scenarios (crisis management).

\end{itemize}

3. Counterfactual \& Reasoning Augmentation: To further enhance complexity, we inject \textbf{counterfactual} turns (e.g., "How would the diagnosis change if the lesions were water-soaked?") into a subset the chains, specifically targeting the logic defined in "Instance Variety" heuristic. 
\subsection{Diverse CoI Scenarios}
\label{sec:coi_traj}
\begin{figure*}[t]
    \centering
    
    % ==================================================================
    % ROW 1: MILD (Early Detection)
    % ==================================================================
    \begin{minipage}[t]{0.28\textwidth}
        \vspace{0pt}
        \centering
        \includegraphics[width=1\linewidth]{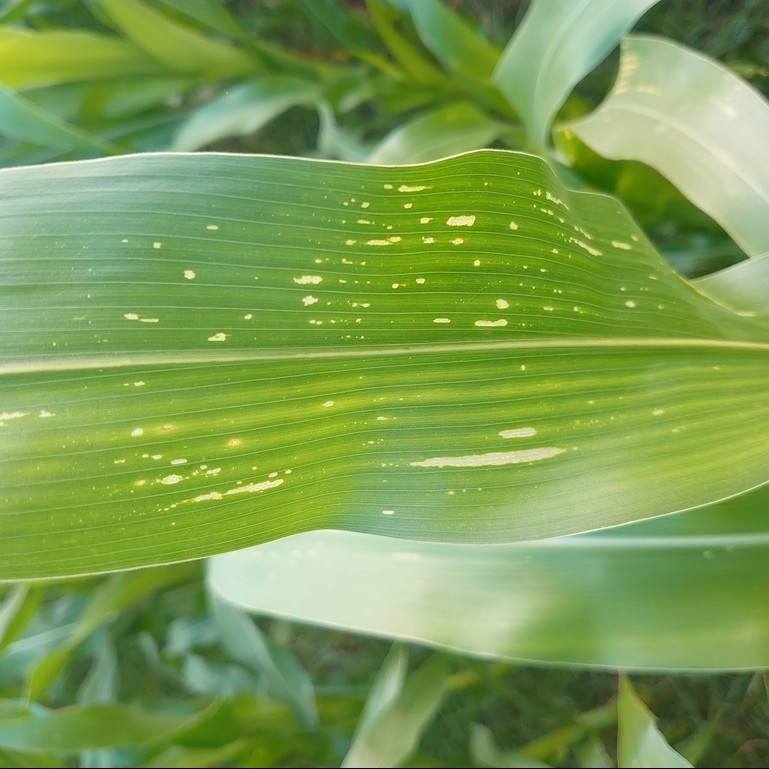}
        \vspace{5pt}
        \textbf{(a) Corn: Mild Infection}
    \end{minipage}
    \hfill
    \begin{minipage}[t]{0.70\textwidth}
        \vspace{0pt}
        \footnotesize
        \textbf{\textcolor{black}{[SEVERITY STATUS: MILD]}}
        
        \vspace{2pt}
        \textbf{Q1:} Can you describe the disease we're seeing here? \\
        \textit{\textbf{A:} We see distinct, discontinuous yellow streaks running parallel to the veins. This is characteristic of early Maize Streak Virus (MSV).} \\
        \textbf{[Cat: Diagnosis | Ground: Yellow streaks, parallel orientation]}
        
        \vspace{2pt}
        \textbf{Q2:} How does this look different from Gray Leaf Spot? \\
        \textit{\textbf{A:} Gray Leaf Spot forms rectangular, gray-to-tan lesions. These are continuous yellow streaks, not necrotic rectangles.} \\
        \textbf{[Cat: Differential Diagnosis | Ground: Lesion shape, color]}
        
        \vspace{2pt}
        \textbf{Q3:} How bad is the infection according to your assessment? \\
        \textit{\textbf{A:} It is mild. The streaks are sparse and the majority of the leaf surface remains green and photosynthetic.} \\
        \textbf{[Cat: Severity Assessment | Ground: Lesion density]}
    \end{minipage}

    \vspace{0.2cm} 
    \hrule
    \vspace{0.2cm}

    % ==================================================================
    % ROW 2: MODERATE (Established Infection)
    % ==================================================================
    \begin{minipage}[t]{0.28\textwidth}
        \vspace{0pt}
        \centering
        \includegraphics[width=1\linewidth]{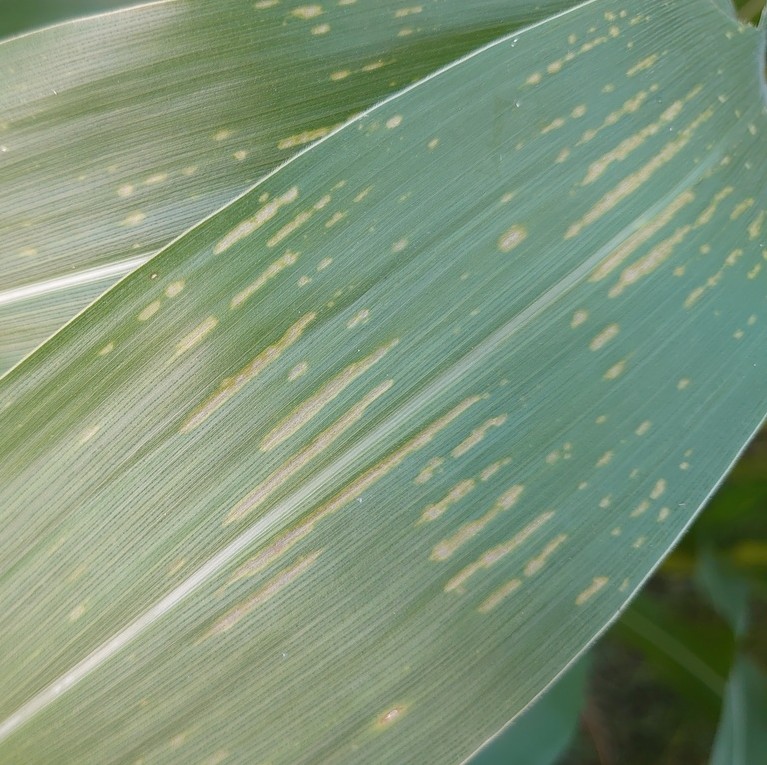}
        \vspace{5pt}
        \textbf{(b) Corn: Moderate Infection}
    \end{minipage}
    \hfill
    \begin{minipage}[t]{0.70\textwidth}
        \vspace{0pt}
        \footnotesize
        \textbf{\textcolor{black}{[SEVERITY STATUS: MODERATE]}}
        
        \vspace{2pt}
        \textbf{Q1:} What's causing these yellow stripes on my corn leaves? \\
        \textit{\textbf{A:} The Maize Streak Virus has progressed. The insect vector (leafhopper) has likely spread the virus, causing these chlorotic bands.} \\
        \textbf{[Cat: Causal Reasoning | Ground: Chlorotic bands]}

        \vspace{2pt}
        \textbf{Q2:} How severe does this look? Is it worth treating? \\
        \textit{\textbf{A:} This is moderate severity. The streaks are coalescing. You must control the leafhopper vector immediately to save the yield.} \\
        \textbf{[Cat: Actionable Severity | Ground: Coalescence]}

        \vspace{2pt}
        \textbf{Q3:} Will these yellow marks get worse if left untreated? \\
        \textit{\textbf{A:} Yes. The chlorosis will expand, reducing the plant's ability to photosynthesize, which will stunt the cob development.} \\
        \textbf{[Cat: Prognosis | Ground: Chlorosis extent]}
    \end{minipage}

    \vspace{0.2cm} 
    \hrule
    \vspace{0.2cm}

    % ==================================================================
    % ROW 3: SEVERE (Critical/Terminal)
    % ==================================================================
    \begin{minipage}[t]{0.28\textwidth}
        \vspace{0pt}
        \centering
        \includegraphics[width=1\linewidth]{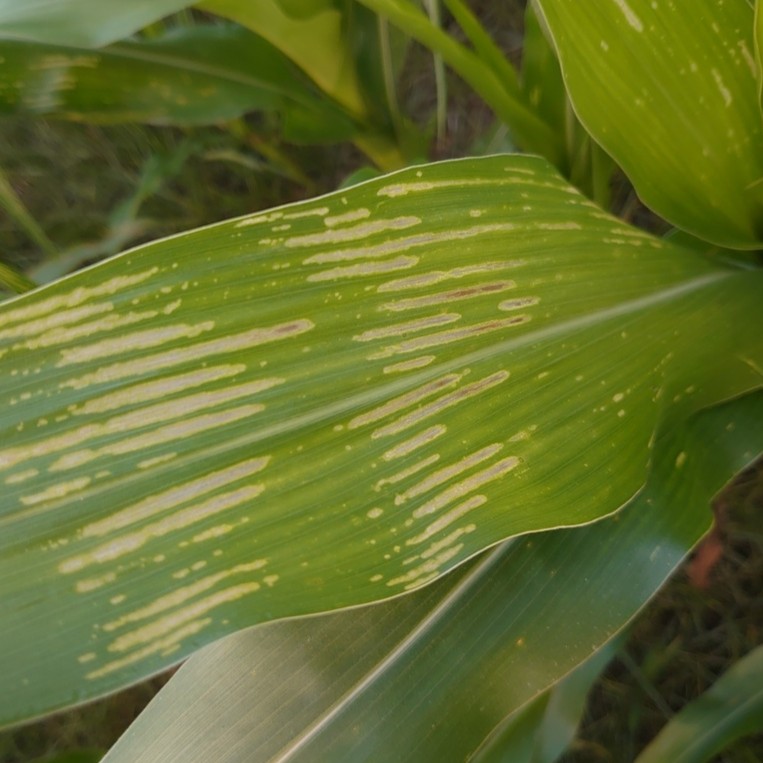}
        \vspace{5pt}
        \textbf{(c) Corn: Severe Infection}
    \end{minipage}
    \hfill
    \begin{minipage}[t]{0.70\textwidth}
        \vspace{0pt}
        \footnotesize
        \textbf{\textcolor{black}{[SEVERITY STATUS: SEVERE]}}
        
        \vspace{2pt}
        \textbf{Q1:} How bad does it look? Is it really severe? \\
        \textit{\textbf{A:} It is severe. The yellow streaking covers almost the entire leaf surface, leaving very little green tissue for photosynthesis.} \\
        \textbf{[Cat: Severity Assessment | Ground: High chlorosis coverage]}

        \vspace{2pt}
        \textbf{Q2:} Do we need to act fast or can we wait? \\
        \textit{\textbf{A:} Urgent action is required, though at this stage, the damage to the current crop's yield is likely irreversible. Focus on preventing spread to neighbors.} \\
        \textbf{[Cat: Urgent Management | Ground: Leaf integrity]}

        \vspace{2pt}
        \textbf{Q3:} Without treating right away, what will happen in the next day or two? \\
        \textit{\textbf{A:} The leaves will likely turn necrotic (brown/dead) and dry out completely, leading to plant death or total yield failure.} \\
        \textbf{[Cat: Consequence Prediction | Ground: Necrosis risk]}
    \end{minipage}

    \caption{\textbf{Diagnostic Reasoning across Disease Severity Stages.} This figure demonstrates how \textbf{PlantInquiryVQA} adapts its questioning strategy as the infection progresses in Maize Streak Virus. 
    \textbf{(a) Mild:} The focus is on \textit{Differential Diagnosis} to distinguish the initial streaks from fungal mimics. 
    \textbf{(b) Moderate:} The inquiry shifts to \textit{Vector Control} and \textit{Prognosis} as the infection becomes established. 
    \textbf{(c) Severe:} The reasoning transitions to \textit{Damage Assessment} and \textit{Salvage}, acknowledging the critical loss of photosynthetic capability.}
    \label{fig:disease_severity}
\end{figure*}

\begin{figure*}[t]
\centering
% ==================================================================
% ROW 1: jackfruit_anthracnose (6 Questions)
% ==================================================================
\begin{minipage}[t]{0.28\textwidth}
\vspace{0pt} 
\centering
\includegraphics[width=1\linewidth]{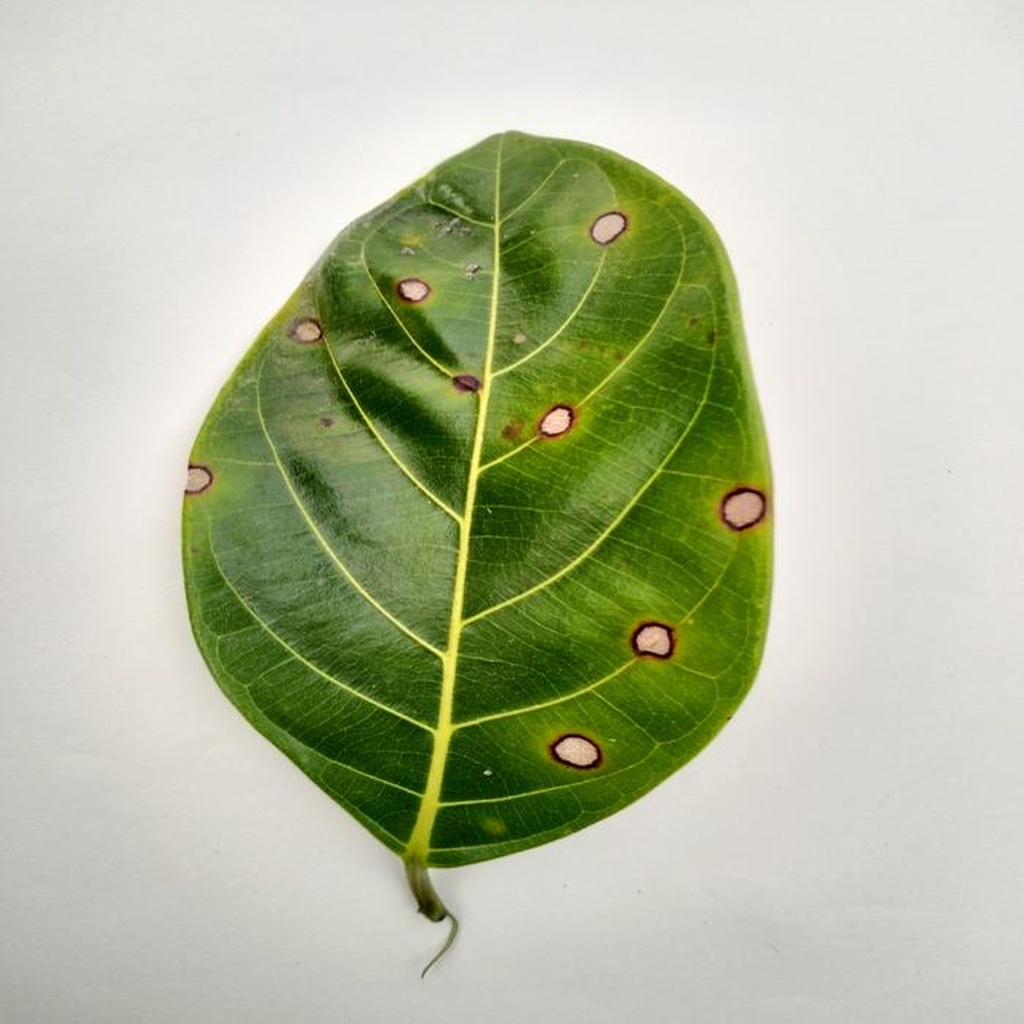} 
\vspace{5pt}
\textbf{(a) Jackfruit: Anthracnose}
\end{minipage}
\hfill
\begin{minipage}[t]{0.70\textwidth}
\vspace{0pt} 
\footnotesize

\textbf{Q1:} How many spots do you see on this jackfruit leaf? \\
\textit{\textbf{A:} Around 10 to 15 spots (moderate). Mostly scattered across midleaf region.} \\
\textbf{[Cat: Severity | Ground: Number of spots, Distribution]}

\vspace{2pt}
\textbf{Q2:} What does the central part of these spots look like? \\
\textit{\textbf{A:} Pale pinkish-white to light tan. Characteristic of anthracnose in jackfruit.} \\
\textbf{[Cat: Lesion Characteristics | Ground: Color of central area]}

\vspace{2pt}
\textbf{Q3:} Are there any signs of the disease getting worse? \\
\textit{\textbf{A:} No, lesions are well-defined with clear red-brown/purple margins. Not merging yet.} \\
\textbf{[Cat: Progression | Ground: Marginal color, Lesion coalescence]}

\vspace{2pt}
\textbf{Q4:} What should we do to manage this anthracnose outbreak? \\
\textit{\textbf{A:} Start by pruning affected leaves/branches. Apply fungicides and ensure air circulation.} \\
\textbf{[Cat: Management | Ground: Treatment]}

\vspace{2pt}
\textbf{Q5:} Could environmental factors contribute to this infection? \\
\textit{\textbf{A:} Yes, high humidity and poor air circulation promote spread.} \\
\textbf{[Cat: Causes | Ground: Environmental conditions]}

\vspace{2pt}
\textbf{Q6:} What changes do you expect to see in the next couple of weeks? \\
\textit{\textbf{A:} If untreated, spots might grow slightly larger but remain scattered.} \\
\textbf{[Cat: Future Outlook | Ground: Expected lesion progression]}
\end{minipage}

\vspace{0.4cm} 
\hrule
\vspace{0.4cm}

% ==================================================================
% ROW 2: grape_anthracnose (6 Questions)
% ==================================================================
\begin{minipage}[t]{0.28\textwidth}
\vspace{0pt}
\centering
\includegraphics[width=1\linewidth]{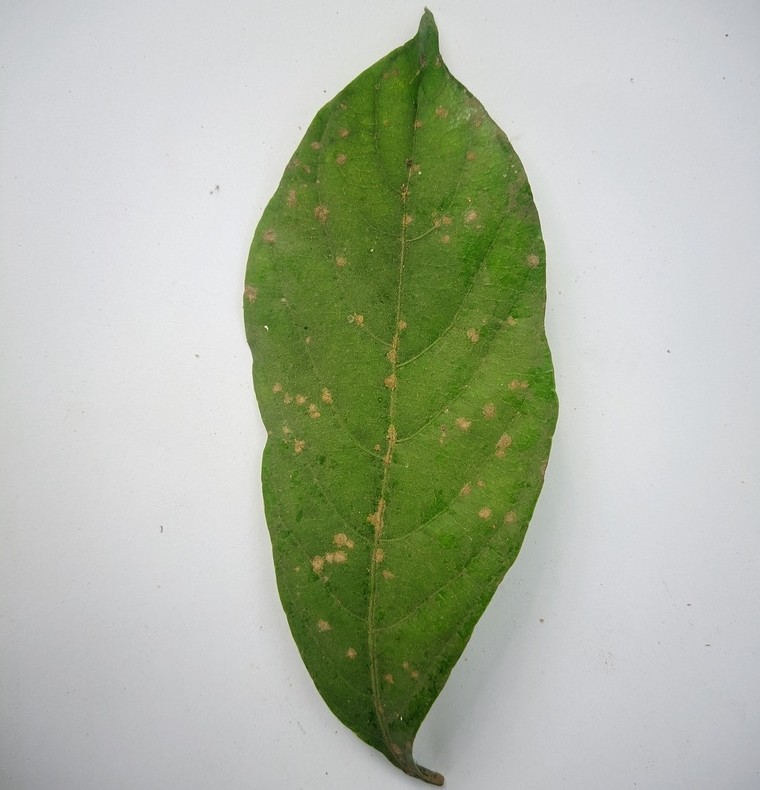}
\vspace{5pt}
\textbf{(b) Grape: Anthracnose}
\end{minipage}
\hfill
\begin{minipage}[t]{0.70\textwidth}
\vspace{0pt}
\footnotesize

\textbf{Q1:} How many spots do you see on this grape leaf? \\
\textit{\textbf{A:} Around 15 to 20 small/medium spots, mostly tan to brown.} \\
\textbf{[Cat: Severity | Ground: Number, Size, Color]}

\vspace{2pt}
\textbf{Q2:} What kind of spots does anthracnose usually make on grape leaves? \\
\textit{\textbf{A:} Small, circular, reddish-brown to black spots. Scattered or clustered near midrib.} \\
\textbf{[Cat: ID | Ground: lesion characteristics, distribution]}

\vspace{2pt}
\textbf{Q3:} Is this infection severe for grapes? \\
\textit{\textbf{A:} Moderate infection (15-20 spots). Not severe yet, but needs attention.} \\
\textbf{[Cat: Severity | Ground: Severity]}

\vspace{2pt}
\textbf{Q4:} Should I start treating these spots right away? \\
\textit{\textbf{A:} Yes, treat immediately to prevent spread. Use recommended fungicides.} \\
\textbf{[Cat: Immediate Management | Ground: Treatment, Management]}

\vspace{2pt}
\textbf{Q5:} What caused these spots on my grape leaves? \\
\textit{\textbf{A:} Likely Colletotrichum spp. fungi. High humidity/poor air favor growth.} \\
\textbf{[Cat: Causes | Ground: Causal agent, Environmental factors]}

\vspace{2pt}
\textbf{Q6:} What changes might we expect in the next couple of weeks? \\
\textit{\textbf{A:} Spots may grow and potentially merge, leading to extensive damage if untreated.} \\
\textbf{[Cat: Future Outlook | Ground: Expected progression, Monitoring]}
\end{minipage}

\caption{\textbf{Cross-species Occurrence of Anthracnose.} The figure illustrates how \textbf{PlantInquiryVQA} adapts its CoI to host-specific manifestations of the same pathogen (\textit{Colletotrichum} spp.). 
\textbf{(a) Jackfruit:} The dialogue identifies the classic "bird's-eye" lesions (pale centers, dark margins) and recommends mechanical intervention (pruning) suitable for tree canopies.
\textbf{(b) Grape:} The dialogue identifies smaller, necrotic tan/brown spots typical of vine infections and prioritizes immediate chemical suppression to prevent rapid merging.
This highlights the model's ability to generate context-aware reasoning that reflects the distinct biological symptoms and management protocols unique to each crop species.}
\label{fig:cross_species}
\end{figure*}

\begin{figure*}[p] % 'p' allows it to take up a dedicated page if needed due to length
\centering

% ==================================================================
% ROW 1: mango_gall_midge (7 Questions)
% ==================================================================
\begin{minipage}[t]{0.28\textwidth}
\vspace{0pt} % Aligns image to top of minipage
\centering
\includegraphics[width=1\linewidth]{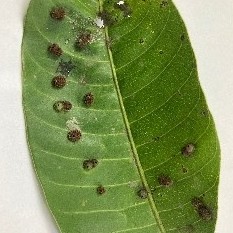} 
\vspace{5pt}
\textbf{(a) Mango: Gall Midge}
\end{minipage}
\hfill
\begin{minipage}[t]{0.70\textwidth}
\vspace{0pt} % Aligns text to top
\footnotesize

\textbf{Q1:} What crop is showing these little brown spots? \\
\textit{\textbf{A:} It's mango trees... spots are small, round, and mostly found on the top side.} \\
\textbf{[Cat: Crop ID | Ground: size, shape, location]}

\vspace{2pt}
\textbf{Q2:} Can you describe the spots on the leaves? \\
\textit{\textbf{A:} The spots are small, brown, round... maybe just a bit bigger than a pinhead.} \\
\textbf{[Cat: Symptoms | Ground: size, color, shape]}

\vspace{2pt}
\textbf{Q3:} How does this compare to other diseases I might see? \\
\textit{\textbf{A:} Different from powdery mildew (no white powder) and bacterial leaf spot (no water-soaked edge).} \\
\textbf{[Cat: Differential Diagnosis | Ground: comparison, edge]}

\vspace{2pt}
\textbf{Q4:} Could you explain what causes these spots? \\
\textit{\textbf{A:} Caused by the gall midge insect laying eggs... larvae form bumps that turn brown.} \\
\textbf{[Cat: Causes | Ground: insect, egg-laying]}

\vspace{2pt}
\textbf{Q5:} Are these spots going to get worse? \\
\textit{\textbf{A:} Damage looks minor. Won’t likely spread much more; tree should recover.} \\
\textbf{[Cat: Prognosis | Ground: spread, severity]}

\vspace{2pt}
\textbf{Q6:} Do you notice anything else unusual besides the spots? \\
\textit{\textbf{A:} Everything else seems normal. No wilting, curled leaves; texture is smooth.} \\
\textbf{[Cat: Additional Cues | Ground: texture, wilting]}

\vspace{2pt}
\textbf{Q7:} What can I do if I want to prevent these spots? \\
\textit{\textbf{A:} Ensure air circulation, keep area clean, avoid overhead watering.} \\
\textbf{[Cat: Prevention | Ground: air\_circulation, cleanliness]}
\end{minipage}

\vspace{0.4cm} 
\hrule
\vspace{0.4cm}

% ==================================================================
% ROW 2: MANGO SOOTY MOLD (6 Questions)
% ==================================================================
\begin{minipage}[t]{0.28\textwidth}
\vspace{0pt}
\centering
\includegraphics[width=1\linewidth]{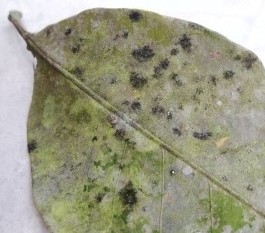}
\vspace{5pt}
\textbf{(b) Mango: Sooty Mold}
\end{minipage}
\hfill
\begin{minipage}[t]{0.70\textwidth}
\vspace{0pt}
\footnotesize

\textbf{Q1:} How widespread is this sooty mould on my mango leaves? \\
\textit{\textbf{A:} Moderately spread across many parts... covering a good portion of surfaces.} \\
\textbf{[Cat: Severity | Ground: overall color, surface texture]}

\vspace{2pt}
\textbf{Q3:} What are the common symptoms of Sooty Mold in mango? \\
\textit{\textbf{A:} Black, fuzzy coating on top side... rubs off easily... associated with honeydew.} \\
\textbf{[Cat: ID | Ground: colors, patterns]}

\vspace{2pt}
\textbf{Q2:} Can you tell me how to control this on my mango trees? \\
\textit{\textbf{A:} Wash leaves with water... apply neem oil/soap (often follows insect infestation).} \\
\textbf{[Cat: Management | Ground: lesion characteristics, distribution]}

\vspace{2pt}
\textbf{Q4:} Why did my mango leaves get these spots? \\
\textit{\textbf{A:} Likely insect infestation producing honeydew... colonized by fungi (e.g., aphids/scale).} \\
\textbf{[Cat: Causes | Ground: lesion characteristics, distribution]}

\vspace{2pt}
\textbf{Q5:} What changes can I expect in the next few days? \\
\textit{\textbf{A:} Mould will continue to grow if conditions remain favorable. Check regularly.} \\
\textbf{[Cat: Future Outlook | Ground: distribution, lesion characteristics]}

\vspace{2pt}
\textbf{Q6:} Is there anything else I need to worry about? \\
\textit{\textbf{A:} Obscures photosynthesis... ensure you manage underlying pests to prevent future issues.} \\
\textbf{[Cat: Impact/Management | Ground: leaf condition]}
\end{minipage}

\caption{\textbf{Multi-disease Occurrence within a Single Crop Species.} The figure demonstrates distinct CoI trajectories for different pathologies affecting the same host (Mango). 
\textbf{(a) Gall Midge:} The dialogue focuses on structural damage (raised bumps), ruling out fungal pathogens via differential diagnosis, and identifying the insect vector. 
\textbf{(b) Sooty Mold:} The dialogue identifies a superficial fungal issue ("rubs off"), links it to secondary insect activity (honeydew), and focuses on cleaning the leaf surface to restore photosynthesis.
This validates that \textbf{PlantInquiryVQA} generates context-specific reasoning chains even when the crop type is identical.}
\label{fig:multi_disease}
\end{figure*}

\begin{figure*}[p]
\centering

% ==================================================================
% ROW 1: DIAGNOSIS (peach_bacterial_spot)
% Focus: Identification & Differential Diagnosis
% ==================================================================
\begin{minipage}[t]{0.28\textwidth}
\vspace{0pt}
\centering
\includegraphics[width=1\linewidth]{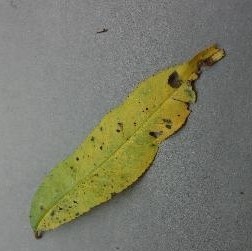}
\vspace{5pt}
\textbf{(a) Peach: Bacterial Spot}
\end{minipage}
\hfill
\begin{minipage}[t]{0.70\textwidth}
\vspace{0pt}
\footnotesize
\textbf{\textcolor{black}{[EPISTEMIC INTENT: DIAGNOSIS]}}

\vspace{2pt}
\textbf{Q1:} What disease is affecting this peach leaf? \\
\textit{\textbf{A:} This is Bacterial Spot (\textit{Xanthomonas campestris}). It appears as small, angular lesions that eventually cause the tissue to drop out.} \\
\textbf{[Cat: Identification | Ground: Angular spots, tissue loss]}

\vspace{2pt}
\textbf{Q2:} How can I tell this isn't fungal "shot hole" disease? \\
\textit{\textbf{A:} Bacterial spots are typically more angular and concentrated near the leaf tip/margins. Fungal shot holes usually have distinct purple halos.} \\
\textbf{[Cat: Differential Diagnosis | Ground: Lesion shape, margins]}

\vspace{2pt}
\textbf{Q3:} What specific visual cue confirms bacteria? \\
\textit{\textbf{A:} The "shot-hole" effect where dead tissue falls out, leaving ragged holes, without the presence of fungal spores.} \\
\textbf{[Cat: Symptomatology | Ground: Holes, ragged edges]}
\end{minipage}

\vspace{0.3cm} 
\hrule
\vspace{0.3cm}

% ==================================================================
% ROW 2: PROGNOSIS (Guava Nutrient Deficiency)
% Focus: Prediction & Future Outlook
% ==================================================================
\begin{minipage}[t]{0.28\textwidth}
\vspace{0pt}
\centering
\includegraphics[width=1\linewidth]{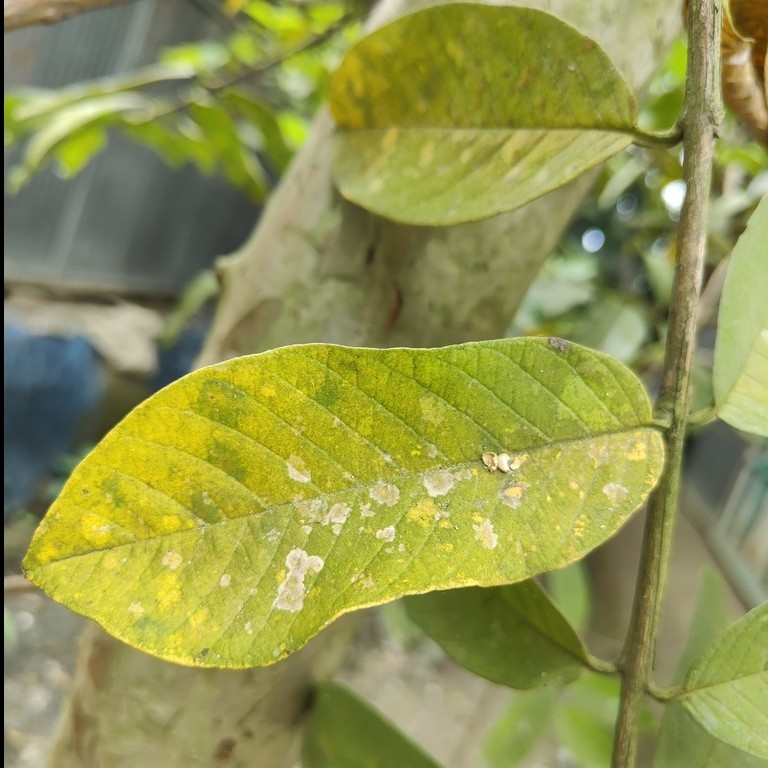}
\vspace{5pt}
\textbf{(b) Guava: Nutrient Deficiency}
\end{minipage}
\hfill
\begin{minipage}[t]{0.70\textwidth}
\vspace{0pt}
\footnotesize
\textbf{\textcolor{black}{[EPISTEMIC INTENT: PROGNOSIS]}}

\vspace{2pt}
\textbf{Q1:} Do you think this will get worse if left untreated? \\
\textit{\textbf{A:} Given it's mild now, if we address nutrient levels, it should improve. Without intervention, it could spread to younger leaves.} \\
\textbf{[Cat: Early Stage Prognosis | Ground: Current severity, distribution]}

\vspace{2pt}
\textbf{Q2:} What changes do you expect in the next few weeks? \\
\textit{\textbf{A:} You might see interveinal chlorosis intensifying. Older leaves may turn completely yellow and drop prematurely.} \\
\textbf{[Cat: Future Outlook | Ground: Chlorosis progression]}

\vspace{2pt}
\textbf{Q3:} Are there any signs of the deficiency getting severe? \\
\textit{\textbf{A:} Not yet. The spotting is scattered and the leaf structure is intact. Severe cases would show necrotic browning.} \\
\textbf{[Cat: Progression | Ground: Leaf integrity, spotting density]}
\end{minipage}

\vspace{0.3cm} 
\hrule
\vspace{0.3cm}

% ==================================================================
% ROW 3: MANAGEMENT (Cauliflower Downy Mildew)
% Focus: Intervention, Treatment, & Prevention
% ==================================================================
\begin{minipage}[t]{0.28\textwidth}
\vspace{0pt}
\centering
\includegraphics[width=1\linewidth]{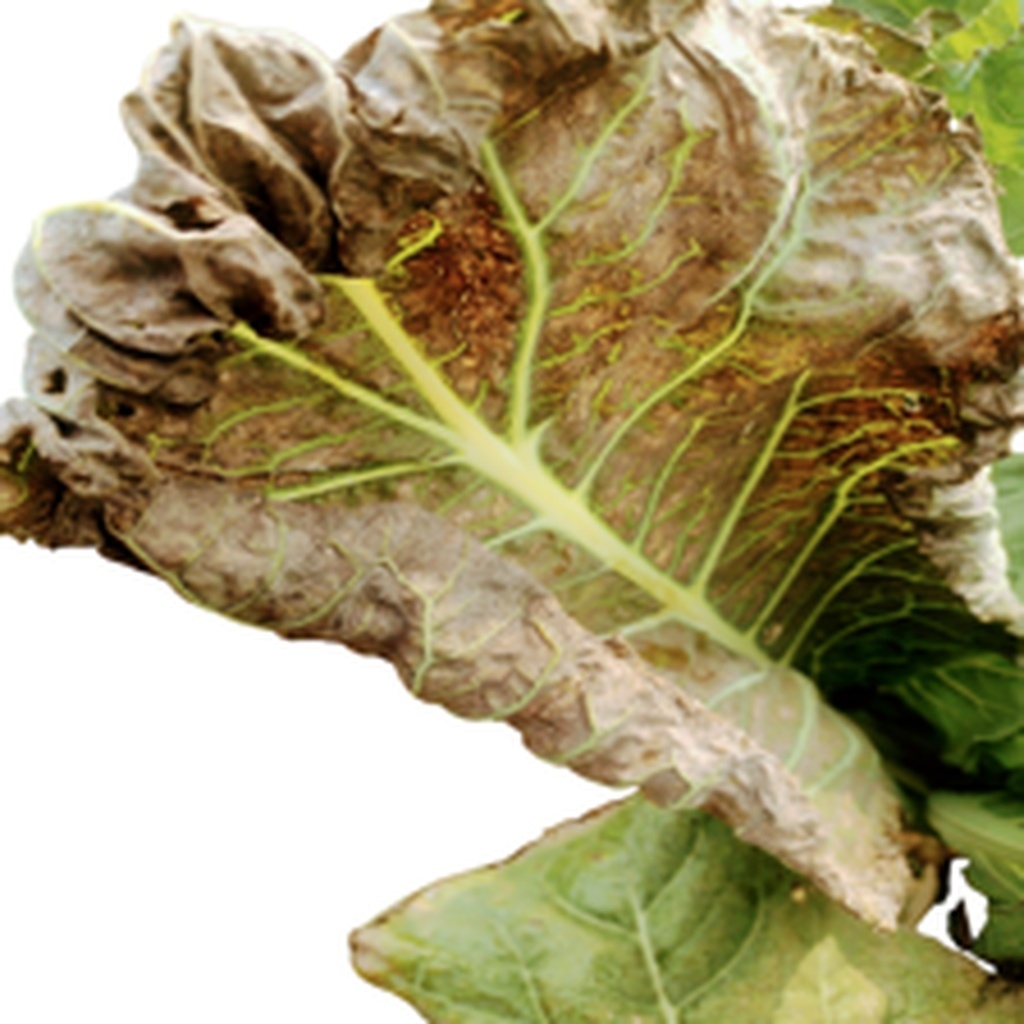}
\vspace{5pt}
\textbf{(c) Cauliflower: Downy Mildew}
\end{minipage}
\hfill
\begin{minipage}[t]{0.70\textwidth}
\vspace{0pt}
\footnotesize
\textbf{\textcolor{black}{[EPISTEMIC INTENT: MANAGEMENT]}}

\vspace{2pt}
\textbf{Q1:} Is it too late to treat these Downy Mildew spots? \\
\textit{\textbf{A:} It’s not too late, but act quickly. Damage is severe/irreversible, but treatment can stop spread to the head.} \\
\textbf{[Cat: Urgent Treatment | Ground: Severity, leaf condition]}

\vspace{2pt}
\textbf{Q2:} What should I have done differently to prevent this? \\
\textit{\textbf{A:} Treated early when small pale spots appeared. Improved drainage and air circulation would have prevented this.} \\
\textbf{[Cat: Prevention | Ground: Lesion characteristics]}

\vspace{2pt}
\textbf{Q3:} If I don't treat this immediately, what happens? \\
\textit{\textbf{A:} The leaf will turn completely brown and die within days. The infection will likely spread to the edible curd.} \\
\textbf{[Cat: Consequence Analysis | Ground: Leaf condition]}
\end{minipage}

\caption{\textbf{Evolution of Epistemic Intent across Disease Severity.} The figure illustrates how the CoI shifts its reasoning goal based on the visual status of the plant. 
\textbf{(a) Diagnosis:} In the early/mild stage (Peach), the focus is on \textit{Identification} and distinguishing symptoms from lookalikes. 
\textbf{(b) Prognosis:} In the mild/chronic stage (Guava), the inquiry shifts to \textit{Predicting} the trajectory of the condition (recovery vs. decline). 
\textbf{(c) Management:} In the severe/terminal stage (Cauliflower), the reasoning pivots entirely to \textit{Intervention}, focusing on salvage, containment, and future prevention strategies.}
\label{fig:epistemic_intent}
\end{figure*}

\begin{figure*}[p]
    \centering
    
    % ==================================================================
    % ROW 1: HEALTHY CONTROL (Cherry)
    % Focus: Validation of Health & Absence of Symptoms
    % ==================================================================
    \begin{minipage}[t]{0.28\textwidth}
        \vspace{0pt}
        \centering
        % Note: Text identifies this as Cherry, though filename was 'blueberry'. 
        \includegraphics[width=1\linewidth]{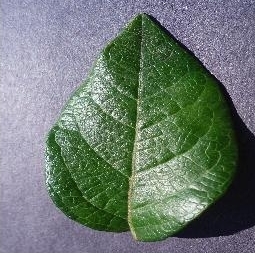} 
        \vspace{5pt}
        \textbf{(a) Cherry: Healthy Control}
    \end{minipage}
    \hfill
    \begin{minipage}[t]{0.70\textwidth}
        \vspace{0pt}
        \footnotesize
        \textbf{{[CONDITION: HEALTHY]}}
        
        \vspace{2pt}
        \textbf{Q1:} What crop is shown in this image? \\
        \textit{\textbf{A:} Cherry.} \\
        \textbf{[Cat: Crop ID]}
        
        \vspace{2pt}
        \textbf{Q2:} Is this leaf healthy or diseased? \\
        \textit{\textbf{A:} Healthy. It shows uniform green color, intact structure, no lesions or spots.} \\
        \textbf{[Cat: Health Status | Ground: Uniform green color, intact structure]}
        
        \vspace{2pt}
        \textbf{Q3:} How does this differ from one affected by fungal diseases? \\
        \textit{\textbf{A:} Fungal diseases cause spots/lesions. The absence of these and presence of uniform green confirm this is healthy.} \\
        \textbf{[Cat: Verification | Ground: Absence of lesions]}
        
        \vspace{2pt}
        \textbf{Q4:} What visual signs would indicate disease on this Cherry leaf? \\
        \textit{\textbf{A:} We would look for discoloration or abnormal growth. This leaf shows none of these indicators.} \\
        \textbf{[Cat: Symptom Indicators | Ground: Intact tissue]}
    \end{minipage}

    \vspace{0.3cm} 
    \hrule
    \vspace{0.3cm}

    % ==================================================================
    % ROW 2: ABIOTIC STRESS / SENESCENCE
    % Focus: Distinguishing Natural Aging from Pathogens
    % ==================================================================
    \begin{minipage}[t]{0.28\textwidth}
        \vspace{0pt}
        \centering
        \includegraphics[width=1\linewidth]{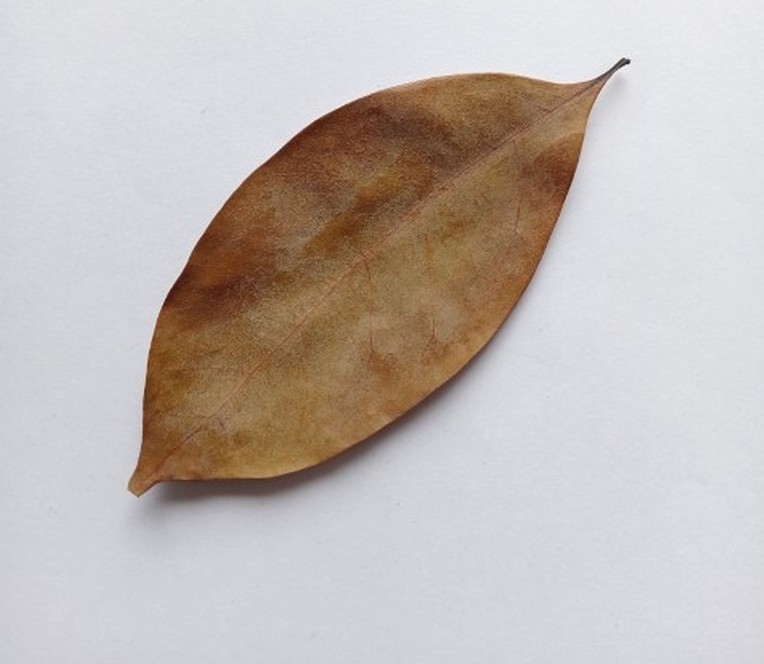}
        \vspace{5pt}
        \textbf{(b) Senescence / Dryness}
    \end{minipage}
    \hfill
    \begin{minipage}[t]{0.70\textwidth}
        \vspace{0pt}
        \footnotesize
        \textbf{{[CONDITION: ABIOTIC STRESS]}}
        
        \vspace{2pt}
        \textbf{Q1:} Can you tell me what’s causing these leaves to turn brown? \\
        \textit{\textbf{A:} Due to senescence or dryness. The whole leaf is brown, dry, and papery.} \\
        \textbf{[Cat: ID | Ground: Uniform brownish-tan, papery texture]}

        \vspace{2pt}
        \textbf{Q2:} How severe is this condition right now? \\
        \textit{\textbf{A:} Quite severe. Entire leaf is dry with no sign of recovery.} \\
        \textbf{[Cat: Severity | Ground: Entire leaf discolored]}

        \vspace{2pt}
        \textbf{Q3:} What might have led to such a severe state? \\
        \textit{\textbf{A:} Likely lack of water, poor soil, or heat stress. Uniform browning suggests environmental stress.} \\
        \textbf{[Cat: Cause | Ground: Uniform surface affection]}

        \vspace{2pt}
        \textbf{Q4:} Do we need to act quickly to stop this from getting worse? \\
        \textit{\textbf{A:} Yes. Without treatment, the leaf will drop off soon.} \\
        \textbf{[Cat: Urgency | Ground: Dryness, no partial vitality]}
    \end{minipage}

    \vspace{0.3cm} 
    \hrule
    \vspace{0.3cm}

    % ==================================================================
    % ROW 3: PEST DAMAGE (Bottle Gourd)
    % Focus: Structural Damage (Holes) vs. Pathogenic Lesions
    % ==================================================================
    \begin{minipage}[t]{0.28\textwidth}
        \vspace{0pt}
        \centering
        \includegraphics[width=1\linewidth]{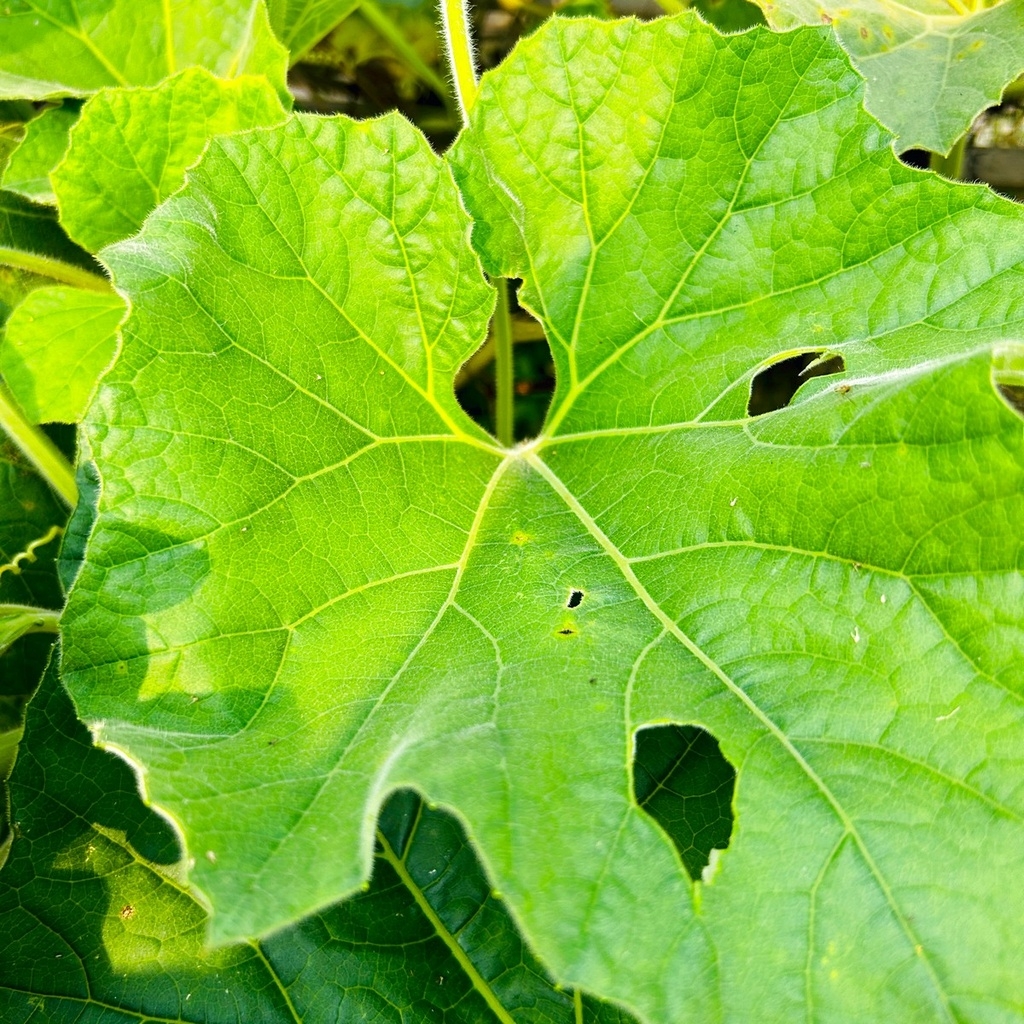}
        \vspace{5pt}
        \textbf{(c) Bottle Gourd: Pest Damage}
    \end{minipage}
    \hfill
    \begin{minipage}[t]{0.70\textwidth}
        \vspace{0pt}
        \footnotesize
        \textbf{{[CONDITION: PEST INFESTATION]}}
        
        \vspace{2pt}
        \textbf{Q1:} How much damage does this look like it's doing? \\
        \textit{\textbf{A:} Moderate damage. Leaves are mostly healthy but have holes and curling edges.} \\
        \textbf{[Cat: Severity | Ground: Leaf condition, holes]}

        \vspace{2pt}
        \textbf{Q2:} Could these holes be from something other than pests, like sunburn? \\
        \textit{\textbf{A:} No, these are clearly pest holes. Sunburn causes scorch marks, not neat-edged openings.} \\
        \textbf{[Cat: Differential Diagnosis | Ground: Holes vs scorch marks]}

        \vspace{2pt}
        \textbf{Q3:} What kind of treatment do you suggest? \\
        \textit{\textbf{A:} Use vegetable-formulated insecticide or organic neem oil.} \\
        \textbf{[Cat: Management | Ground: Lesion characteristics]}

        \vspace{2pt}
        \textbf{Q4:} Do you think the damage will get worse if I don’t do anything? \\
        \textit{\textbf{A:} Yes, larger lesions may grow and affect more surface area.} \\
        \textbf{[Cat: Prognosis | Ground: Lesion distribution]}

        \vspace{2pt}
        \textbf{Q5:} How can I prevent this from happening again? \\
        \textit{\textbf{A:} Crop rotation, row covers, and removing fallen debris.} \\
        \textbf{[Cat: Prevention | Ground: Causes]}
    \end{minipage}

    \caption{\textbf{Beyond Pathogenic Disease: Healthy, Abiotic, and Pest Conditions.} This figure illustrates the dataset's coverage of diverse plant health states. 
    \textbf{(a) Healthy Control:} The model validates health by citing "uniform green color" and the absence of lesions. 
    \textbf{(b) Senescence:} The inquiry identifies abiotic stress (aging/dryness) based on global uniform browning and papery texture, distinguishing it from focal infection. 
    \textbf{(c) Pest Damage:} The reasoning chain differentiates physical damage (holes) from pathogenic spots and recommends insect-specific treatments (Neem oil) rather than fungicides.}
    \label{fig:diverse_conditions}
\end{figure*}

The \textbf{CoI} trajectories across all 12 distinct scenarios are shown in Figure \ref{fig:disease_severity}, Figure \ref{fig:cross_species}, Figure \ref{fig:multi_disease}, Figure \ref{fig:epistemic_intent} and Figure \ref{fig:diverse_conditions}.
% Ensure this block is NOT inside another \begin{figure} ... \end{figure}

\subsection{Semantic Accuracy Evolution per Inquiry Step}
\label{sec:semantic_evolution}

\begin{figure*}[t]
  \includegraphics[width = 0.95\linewidth]{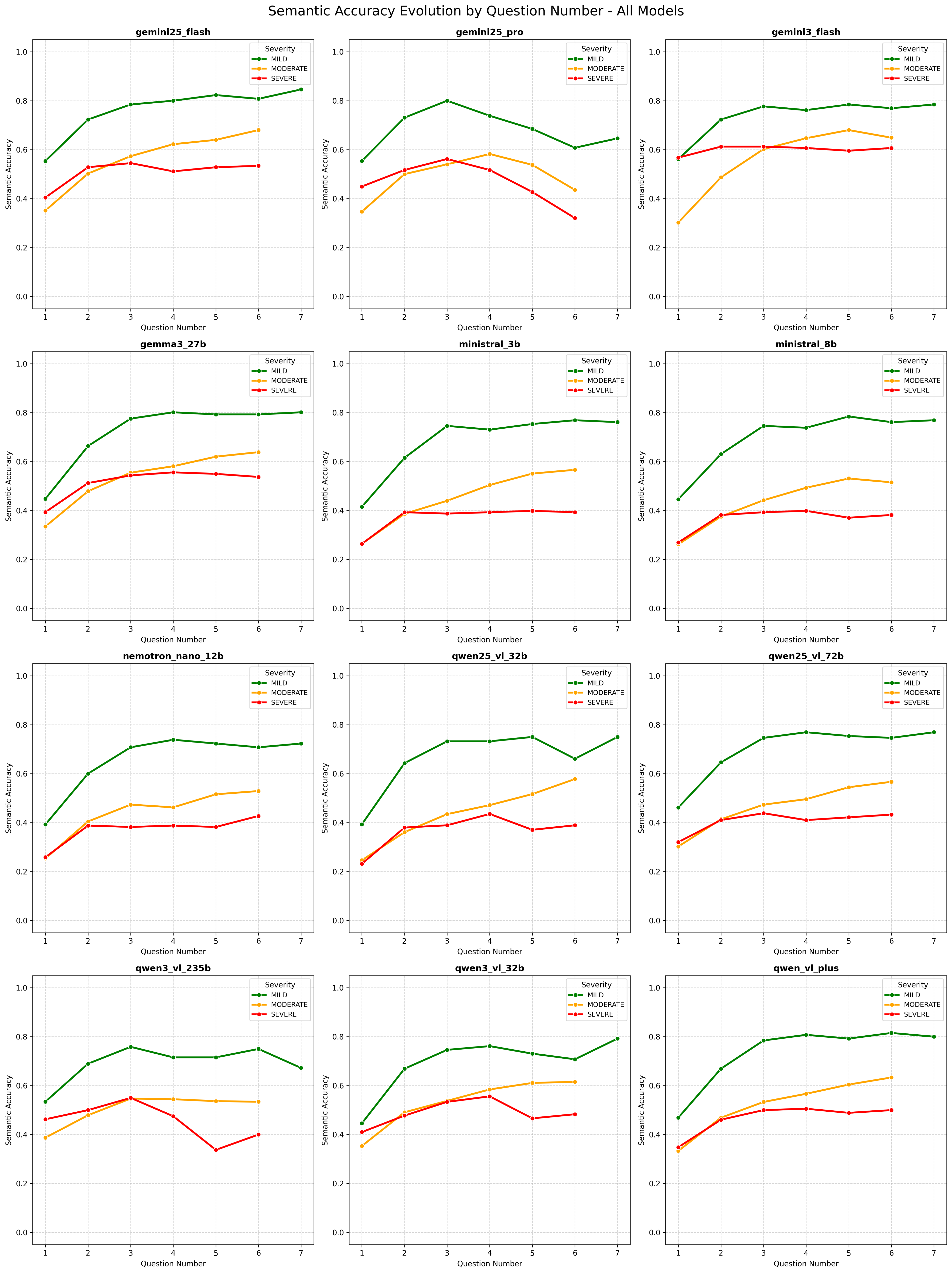} 
  \caption{\textbf{Semantic Accuracy Evolution across the Chain-of-Inquiry Trajectory}. The figure illustrates the layer-wise diagnostic accuracy improvement for all 12 evaluated models as they progress through the 7-step diagnostic inquiry. Green lines indicate Mild infection, showing the strongest positive trajectory, while Red lines (Severe) indicate lower baselines and higher volatility. We observe a consistent positive reasoning trajectory, where accuracy improves with each subsequent question, validating the hypothesis that structured inquiry refines diagnostic precision over time. Notably, the performance is stratified by disease severity: models consistently achieve the highest accuracy on Mild cases (green lines), where visual symptoms are distinct, but struggle with Severe cases (red lines), where extensive tissue necrosis often obscures the discriminative features required for accurate grounding.}
  \label{fig:semantic_evolution}
\end{figure*}

Figure \ref{fig:semantic_evolution} presents a comprehensive layer-wise analysis of semantic accuracy evolution across the seven-step diagnostic trajectory for all 12 evaluated models. By stratifying performance across three disease severity levels—\textit{Mild} (Green), \textit{Moderate} (Orange), and \textit{Severe} (Red)—we observe a distinct divergence in reasoning stability. High-performing models generally exhibit a monotonic increase in semantic alignment as the inquiry progresses, confirming the CoI framework's ability to refine diagnostic grounding through iterative evidence accumulation. However, this positive trajectory is notably dampened in severe cases, where extensive tissue necrosis introduces visual ambiguity that limits the efficacy of the refinement process compared to the linear gains observed in mild infection phenotypes.

% \begin{figure*}[h]
%     \centering
%     \includegraphics[width=1.0\textwidth]{all_models_semantic_evolution.jpg}
%     \caption{\textbf{Semantic Accuracy Evolution by Question Step.} The plots track the semantic similarity score (y-axis) across the 7-step Chain-of-Inquiry (x-axis) for all evaluated models. Green lines indicate Mild infection, showing the strongest positive trajectory, while Red lines (Severe) indicate lower baselines and higher volatility.}
%     \label{fig:semantic_evolution}
% \end{figure*}

\begin{figure*}[t]
  \includegraphics[width = 0.95\linewidth]{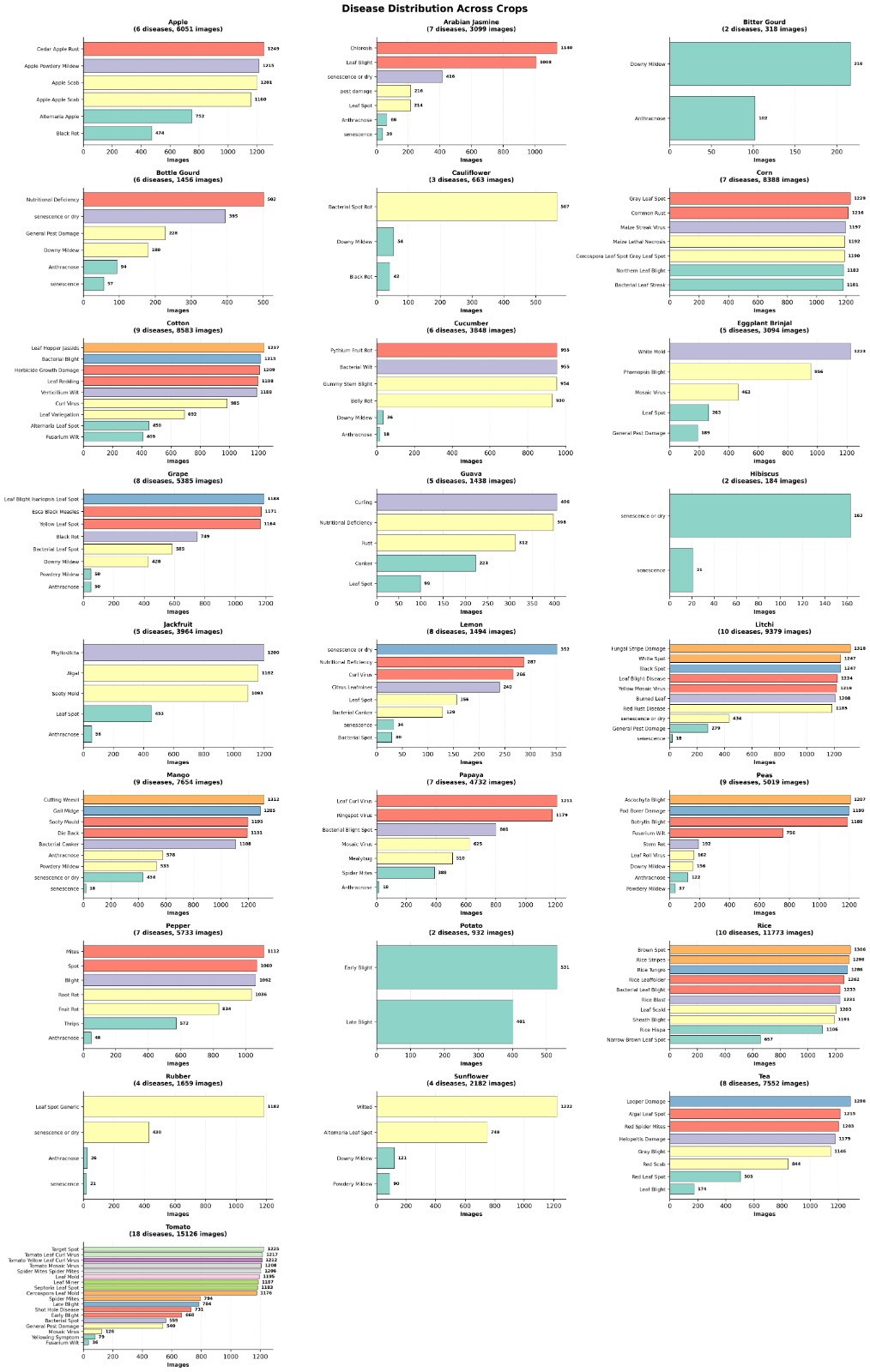} 
  \caption {A comprehensive analysis of diverse disease distribution across crops species of the final PlantInquiryVqa dataset}
  \label{tot_crop_disease}
\end{figure*}

\end{document}